\documentclass[lettersize,journal]{IEEEtran}
\usepackage{amsmath,amsfonts}
\usepackage{algorithmic}
\usepackage{algorithm}
\usepackage{array}
\usepackage{textcomp}
\usepackage{stfloats}
\usepackage{url}
\usepackage{verbatim}
\usepackage{graphicx}
\usepackage{cite}
\usepackage{booktabs}
\usepackage{caption}
\usepackage{subcaption}
\usepackage{multirow}
\usepackage[table]{xcolor}
\usepackage{hyperref}
\usepackage{makecell} 

\newcommand{\eg}{\emph{e.g.}}
\newcommand{\ie}{\emph{i.e.}} 
\newcommand{\Rebuttal}[1]{\textcolor{black}{#1}}

\newcommand{\Rebuttalmr}[1]{\textcolor{black}{#1}}

\newtheorem{assumption}{\bf Assumption}
\newtheorem{proposition}{\bf Proposition}

\begin{document}

\title{Disentangled Generative \\ Graph Representation Learning}

\author{
	Xinyue~Hu\href{https://orcid.org/0009-0007-2603-4670}{\includegraphics[scale=0.06,bb=0 -60 100 200]{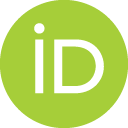}},
	Zhibin~Duan\href{https://orcid.org/0009-0000-3353-9975}{\includegraphics[scale=0.06,bb=0 -60 100 200]{new_fig/ORCIDiD.png}},
	Xinyang~Liu,
	Yuxin~Li\href{https://orcid.org/0000-0002-5935-0684}{\includegraphics[scale=0.06,bb=0 -60 100 200]{new_fig/ORCIDiD.png}},
	Bo~Chen\href{https://orcid.org/0000-0001-5151-9388}{\includegraphics[scale=0.06,bb=0 -60 100 200]{new_fig/ORCIDiD.png}},~\IEEEmembership{Senior~Member,~IEEE},\\
	Chaojie~Wang\href{https://orcid.org/0000-0002-7644-7621}{\includegraphics[scale=0.06,bb=0 -60 100 200]{new_fig/ORCIDiD.png}},
	Yilin He,
	Hongwei~Liu\href{https://orcid.org/0000-0003-4046-163X}{\includegraphics[scale=0.06,bb=0 -60 100 200]{new_fig/ORCIDiD.png}},~\IEEEmembership{Member,~IEEE},
	Xuefei~Cao,~\IEEEmembership{Member,~IEEE},
	and Mingyuan~Zhou
	\thanks{This work was supported in part by the National Natural
		Science Foundation of China under Grant 62576266 and U21B2006; in part by the Fundamental Research Funds for the Central Universities QTZX24003 and QTZX23018; in part by the 111 Project under Grant B18039. (Xinyue Hu and Zhibin Duan contributed equally to this work. \textit{Corresponding authors: Bo Chen and Hongwei Liu.})}
	\thanks{Xinyue Hu, Zhibin Duan, Yuxin Li, Bo Chen, Chaojie Wang and Hongwei Liu are with the National Key Laboratory of Radar Signal Processing, Xidian University, Xi’an, 710071, China; and also with the Institute of Information Sensing, Xidian University, Xi’an 710071, China (e-mail: xinyuehu122@gmail.com; xd\_zhibin@163.com; liyuxin@xidian.edu.cn; bchen@mail.xidian.edu.cn; xd\_silly@163.com; hwliu@xidian.edu.cn).}
	\thanks{Xuefei Cao is with School of Cyber Science, Xidian University, Xi'an, Shaanxi 710071, China (email: xfcao@xidian.edu.cn).}
	\thanks{Xinyang Liu, Yilin He and Mingyuan Zhou is with McCombs School of Business, The University of Texas at Austin, TX 78712, USA (email: xinyangATK@gmail.com; yilin.he@mccombs.utexas.edu; mingyuan.zhou@mccombs.utexas.edu).}%
	\thanks{Manuscript received April 15, 2025; revised October 27, 2025; revised January 26, 2026; accepted March 23, 2026.}
}




\markboth{Journal of \LaTeX\ Class Files,~Vol.~14, No.~8, August~2021}%
{Shell \MakeLowercase{\textit{et al.}}: A Sample Article Using IEEEtran.cls for IEEE Journals}


\maketitle
\begin{abstract}
Generative graph models have recently emerged as a powerful paradigm in self-supervised learning for graph representation tasks, demonstrating remarkable potential in capturing complex structural and feature information. Despite their promise, most existing Generative Graph Representation Learning (GRL) methods treat the entire graph as a monolithic entity, employing random masking strategies across the entire structure. This approach often neglects the intricate interplay between different substructures of the graph, leading to entangled representations. Such entanglement can reduce the robustness of learned models and limit their ability to produce reliable outputs.
To address these limitations, this paper presents \textit{DiGGR} (\textit{Di}sentangled \textit{G}enerative \textit{G}raph \textit{R}epresentation Learning), a novel self-supervised learning framework designed to disentangle latent factors in graph representations. DiGGR introduces a disentanglement-driven approach to guide graph mask modeling, enabling the framework to learn distinct latent factors that better capture the underlying structure of graphs. 
By incorporating these latent factors into an end-to-end joint learning process, DiGGR enhances the clarity and structural alignment of the learned representations, while also improving their robustness and generalization across tasks.
The effectiveness of DiGGR is validated through extensive experiments on 15 public datasets spanning three distinct graph learning tasks. The results consistently demonstrate that DiGGR outperforms a wide range of previous self-supervised methods, highlighting its superior ability to learn meaningful, disentangled representations. These findings underscore the potential of DiGGR to advance the state of the art in Generative GRL research.

\end{abstract}


\section{Introduction}\label{intro}
\IEEEPARstart{S}{elf}-supervised learning (SSL) has received attention due to its appealing capacity for learning data representation without label supervision.
While contrastive SSL approaches are becoming increasingly utilized on images  \cite{chen2020simple} and graphs \cite{you2020graph}, generative SSL has emerged as a complementary and increasingly significant direction, fueled by influential advancements such as BERT for natural language processing \cite{devlin2018bert}, BEiT for vision tasks \cite{bao2021beit}, and MAE for image representation learning \cite{he2022masked}. These generative approaches focus on learning robust and context-aware representations by reconstructing masked or corrupted input data, providing a powerful alternative to contrastive methods.
Along this line, generative SSL methods have been extended to graph-structured data, leading to the emergence of graph masked autoencoders (GMAE). 
Generally, the fundamental concept of GMAE \cite{tan2022mgae} is to utilize an autoencoder architecture to reconstruct input node features, structure, or both, which are randomly masked before the encoding step.
The model is tasked with reconstructing these masked elements during the decoding phase, enabling it to learn representations that capture both local node-level features and the global topological structures of the graph. This approach has achieved considerable success across a wide range of applications, including node classification \cite{kipf2016semi} and graph classification \cite{xu2018powerful}. Furthermore, it has spurred the development of numerous tailored GMAE frameworks \cite{hou2022graphmae, tu2023rare, tian2023heterogeneous}, each introducing innovative architectural designs or training strategies to improve performance on diverse graph analysis tasks.

Despite their achievements, most GMAE approaches typically treat the entire graph as a holistic entity, neglecting the intricate latent structures inherent within. 
In detail, these methods typically utilize a uniform random masking strategy applied indiscriminately across the graph's nodes and edges.
This simplistic approach tends to result in node representations that encapsulate their neighborhood as a perceptual whole, disregarding the nuanced distinctions between different parts of the neighborhood \cite{ma2019disentangled, li2021disentangled, mo2023disentangled, gao2024hypergraph}.
For example, in a social network $G$, individual $n$ is a member of both a mathematics group and several sports interest groups. Due to the diversity of these different communities, they may exhibit different characteristics when interacting with members from various communities. Specifically, the information about the mathematics group may be related to their professional research, while the information about sports clubs may be associated with their hobbies. 
Existing GMAE methods, by not disentangling these heterogeneous factors, amalgamate these distinct sources into a single representation. As a result, the learned representation fails to capture the diversity of interactions and attributes, often entangling unrelated factors.
This limitation not only diminishes the robustness of the representations by making them susceptible to noise, but also compromises the discriminability of the learned representations, as they may be easily influenced by irrelevant factors \cite{hou2022graphmae}.

\begin{figure*}[t]
\centering
\includegraphics[width=18cm]{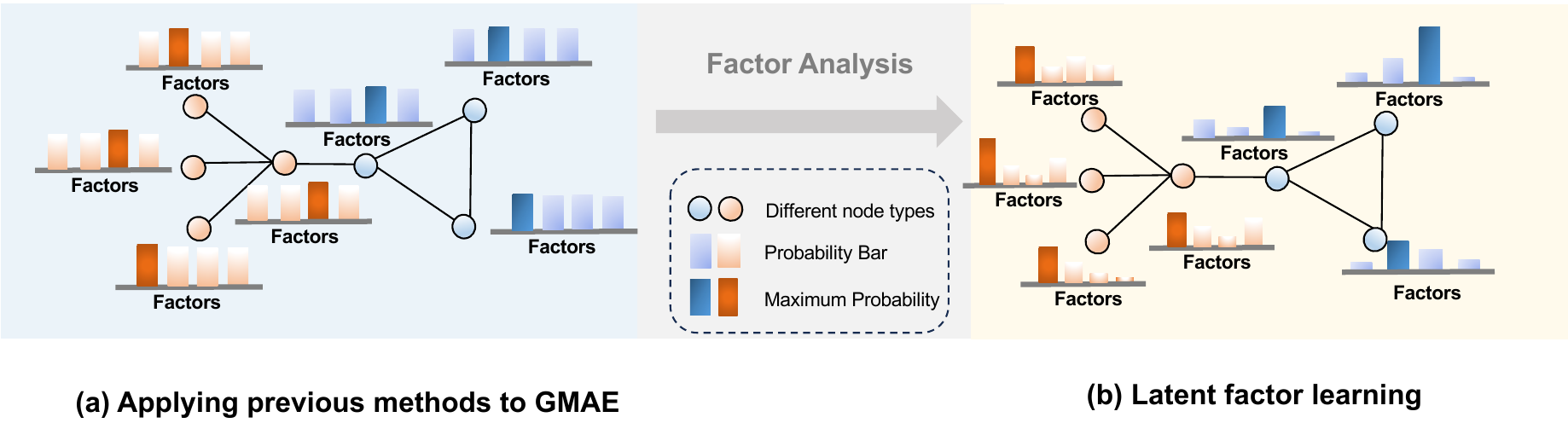} 
\caption{ 
The number of latent factors is set to 4. In (a), \Rebuttal{directly applying conventional factor learning to GMAE results in poor convergence, with similar node-factor affiliation probabilities causing misassignment.} Fig. 1(b) shows that the probabilities of node-factor affiliation are more discriminative, correctly categorizing nodes of the same type into the same latent group.
}
\label{fig: motivation}
\end{figure*} 

To alleviate the challenge described above, disentangled graph representation learning has gained increasing attention as a promising direction for capturing the complex and diverse latent factors underlying graph data \cite{bengio2013representation, li2021disentangled, ma2019disentangled, mo2023disentangled, xiao2022decoupled}. The primary objective of disentangled graph representation learning is to extract representations that effectively separate and encode the distinct factors of variation within a graph, thereby enhancing robustness and improving generalization to diverse structural and attribute patterns.
Specifically, most existing approaches achieve this by incorporating a self-clustering module designed to identify latent factors. These modules typically rely on measuring the similarity between node representations and latent factor prototypes. Through this clustering mechanism, nodes are grouped based on shared latent characteristics, allowing the model to produce factor-specific representations that encapsulate the underlying structure and relationships within the graph. 
By leveraging these acquired latent factors, these models adeptly capture factor-wise graph representations, effectively encapsulating the latent structure of the graph.
However, despite the advancements in disentangled representation learning, its application to generative graph representation learning methods, such as GMAE, remains relatively unexplored. 
A key challenge lies in the difficulty of achieving convergence in the latent factor detection module under the generative training target, thus presenting obstacles in practical implementation.
As shown in Fig. \ref{fig: motivation}(a), directly applying conventional factor learning methods to GMAE would make the factor learning module difficult to converge, resulting in 
poorly differentiated probabilities of node-factor affiliation, causing nodes with similar properties to be erroneously assigned to different latent factor groups. This misalignment undermines the effectiveness of disentangled representations, limiting their utility in capturing the nuanced and multifaceted structures of real-world graphs.

To overcome these limitations, we introduce \textbf{Di}sentangled \textbf{G}enerative \textbf{G}raph \textbf{R}epresentation Learning (\textbf{DiGGR}), a novel generative self-supervised framework designed to capture and utilize latent disentangled factors for enhanced graph representation. 
Generally speaking, DiGGR operates by learning to generate graph structures from latent disentangled factors $z$, leveraging these factors to guide the reconstruction of masked graph components while maintaining end-to-end joint optimization. This dual capability allows the model to effectively harness the inherent heterogeneity of graph data. Specifically, the framework incorporates the following innovative modules:
$i)$ Probabilistic Latent Factor Learning Module: This module is designed to capture the diverse underlying factors that govern node and edge formations within a graph. By modeling the generation process probabilistically, it enables the decomposition of the graph into multiple disentangled subgraphs, each representing a distinct latent factor, thus enhancing structural alignment and robustness of the learned representations. 
$ii)$ Factor-wise Self-Supervised Learning Framework: To further enhance the representation quality, the framework employs a unique masking and reconstruction strategy for each disentangled subgraph. This factor-specific masking approach enables the model to learn representations that are both detailed and robust, tailored to the unique characteristics of each factor. By doing so, the framework effectively integrates local and global information, leading to a richer and more nuanced representation.
Evaluation shows that the proposed framework can achieve significant performance enhancement on various node and graph classification benchmarks.
{The main contributions of this paper can be summarized as follows:} 
\begin{itemize}
\item {
\Rebuttal{We propose \textbf{DiGGR} (\textbf{Di}sentangled \textbf{G}enerative \textbf{G}raph \textbf{R}epresentation Learning), which introduces a probabilistic factorization mechanism to learn unsupervised, disentangled representations within a generative graph-reconstruction framework.}
}

\item{\Rebuttal{We introduce a factor-global collaborative masking paradigm, where factor-specific local masking and graph-level global reconstruction work in concert to inject disentangled information into the learned representations. Further, we take theoretical analysis to verify its disentangled properties.}}

\item{Empirical results show that the proposed DiGGR outperforms many previous self-supervised methods in various node- and graph-level classification tasks.}
\end{itemize}

\section{Related works}

\subsection{Graph Self-Supervised Learning}
\label{subsec_2.2: GSSL}
Graph Self-Supervised Learning (SSL) has achieved remarkable advancements in addressing the challenge of label scarcity in real-world network data. These advancements can be broadly categorized into two major paradigms: contrastive learning and generative learning, each offering unique perspectives and methodologies for representation learning.
Graph contrastive learning is largely grounded in mutual information maximization \cite{DBLP:conf/iclr/HjelmFLGBTB19, DBLP:conf/iclr/VelickovicFHLBH19}, aligning representations of the same instance across augmented views and thus becoming a dominant paradigm for graph SSL. Its performance mainly depends on augmentation quality, which typically falls into three categories: (1) feature perturbation \cite{hu2019strategies, zhu2020deep, DBLP:conf/iclr/VelickovicFHLBH19}, (2) structure/proximity modification \cite{hassani2020contrastive, you2020graph}, and (3) graph sampling \cite{qiu2020gcc}. Despite their success, contrastive methods often require carefully designed pretext tasks and augmentations, involving substantial trial-and-error and added computational cost. Recent attempts reduce this reliance by avoiding explicit augmentations in feature space \cite{DBLP:conf/www/XiaWCHL22, DBLP:journals/tnn/LiCZLHW24}, but they still construct multiple views, leaving room to further minimize dependence on augmentation strategies.
Another major paradigm, generative SSL, learns representations by reconstructing missing parts of the input. Early methods such as Graph Autoencoders (GAE) and Variational Graph Autoencoders (VGAE) follow the autoencoding framework \cite{hinton1993autoencoders, kipf2016variational}, encoding nodes into low-dimensional embeddings and reconstructing graph structure, which is conceptually simpler and easier to integrate than contrastive approaches. More recently, Graph Masked Autoencoders (GMAEs) have advanced generative SSL by mitigating limitations of traditional generative methods that can overemphasize neighborhood information while overlooking finer structural details \cite{hassani2020contrastive, DBLP:conf/iclr/VelickovicFHLBH19, cao2024immense}. Through masking graph structure \cite{li2023s}, node attributes \cite{hou2022graphmae}, or both \cite{tian2023heterogeneous}, GMAEs improve representation granularity and robustness. However, most GMAEs adopt random masking over the whole graph, treating a node’s neighborhood as a single entity and ignoring distinctions within it. To address this, we propose a disentangled generative representation learning framework that captures the unique characteristics of different structural components in a node’s neighborhood.

\subsection{Disentangled Graph Learning}
Disentangled representation learning aims to identify and isolate the underlying explanatory factors in data \cite{bengio2013representation}. While most prior work is developed in computer vision \cite{higgins2017beta, jiang2020psgan}, recent studies have extended disentanglement to graph-structured data \cite{li2021disentangled, ma2019disentangled, mercatali2022symmetry, mo2023disentangled}. For instance, \cite{ma2019disentangled} uses attention to separate latent factors, producing node embeddings that capture semantic dimensions, and \cite{li2021disentangled} learns disentangled graph-level representations via self-supervision so that factorized representations capture complementary information. Other methods \cite{li2022disentangled, wang2022deep, wang2020disentangled} also report gains on downstream tasks. However, many of these approaches \cite{ma2019disentangled, li2021disentangled, li2022disentangled, wang2020disentangled} learn factors in a prototype-based manner by mapping node representations to simplex vectors with a softmax, which often leads to latent-factor convergence issues when applied to GMAE. Moreover, they typically emphasize task-related signals while under-utilizing graph structural information. To address these limitations, we propose DiGGR, which models latent factors with a Gamma distribution rather than point vectors and incorporates both structural information and task-related information. \Rebuttal{Related probabilistic graph models include EPM \cite{zhou2015infinite} for community discovery and VEPM \cite{he2022variational} for supervised representation learning. Although sharing blocks such as Gamma/Weibull distributions, they differ in objective and formulation: EPM focuses on generation and VEPM relies on explicit supervision, whereas DiGGR uses probabilistic factorization as a structural inductive bias within a self-supervised masked autoencoding framework. In contrast to non-graph probabilistic models \cite{huenhancing, zhang2018whai}, we explicitly exploit distributional properties.} Leveraging Gamma-induced sparsity and graph structure, DiGGR extracts more distinguishable factors and enables end-to-end self-supervised disentangled generative graph representation learning.



\section{Proposed Method}
\begin{figure*}[t]
\centering
\includegraphics[width=18cm]{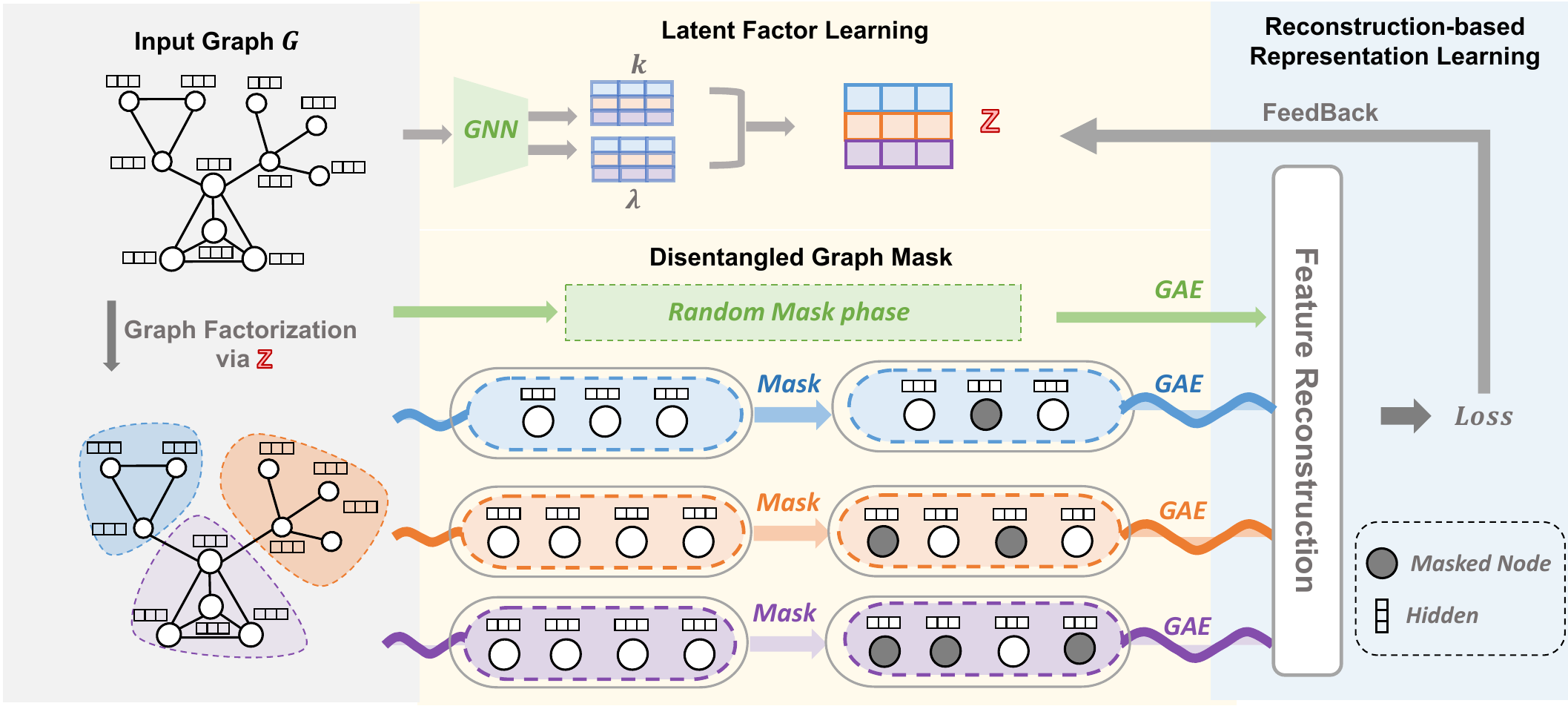} 

\caption{
\Rebuttalmr{Overview of DiGGR’s computation graph. The input passes through three modules (Sections \ref{subsec_3.2: latent_factor} and \ref{subsec_3.3: disentangled GMAE}): Latent Factor Learning, Graph Factorization, and Disentangled Graph Mask Autoencoder. Latent Factor Learning extracts the latent factor $z$ from the input graph, and Graph Factorization uses $z$ to produce factorized subgraphs where nodes exchange more information with strongly interacting neighbors. During disentangled masking, we mask each factorized subgraph individually to enhance the disentanglement of the learned node representations.}
}
\label{fig: model}
\end{figure*}

In this section, we propose \textbf{DiGGR} (\textbf{Di}sentangled \textbf{G}enerative \textbf{G}raph \textbf{R}epresentation Learning) for self-supervised graph representation learning with mask modeling. 
The framework, illustrated in Figure~\ref{fig: model}, consists of three main components: Latent Factor Learning (Section~\ref{subsec_3.2: latent_factor}), Graph Factorization, which encompasses both node and edge factorization (also in Section~\ref{subsec_3.2: latent_factor}), and the Disentangled Graph Masked Autoencoder (Section~\ref{subsec_3.3: disentangled GMAE}).
Before elaborating on them, we first show some notations.

\subsection{Preliminaries}


A graph \( G \) can be formalized as a tuple \( \mathcal{G} = \{ V, A, X \} \), where \( V \) represents the set of nodes, \( A \) denotes the set of edges, and \( X \) corresponds to the feature matrix of the nodes. Specifically, let \( \lvert V \rvert = N \) represent the number of nodes in the graph, \( \lvert A \rvert = M \) the number of edges, and \( X \in \mathbb{R}^{N \times L} \) a feature matrix where each node is associated with an \( L \)-dimensional feature vector. The graph’s topology, which defines the structure of connections between the nodes, is captured in its adjacency matrix \( A \in \mathbb{R}^{N \times N} \), where each element \( A_{ij} \) indicates the presence or absence of an edge between nodes \( i \) and \( j \).

Additionally, we introduce a latent disentangled factor matrix \( z \in \mathbb{R}^{N \times K} \), where \( K \) denotes the number of predefined factors. This factor matrix represents the hidden structure of the graph, which will later guide the mask modeling process. To achieve this, we utilize a probabilistic graph generation model that aims to decompose the graph into latent factors, each corresponding to a distinct aspect of the graph’s structure.

Given the graph \( G \), we decompose it into \( \{ G_1, G_2, \dots, G_K \} \), where each \( G_k \) is a factor-specific subgraph. Each subgraph \( G_k \) consists of its own specific edge set \( A^{(k)} \), node set \( V^{(k)} \), and node feature matrix \( X^{(k)} \), which captures the part of the graph that is associated with the factor \( k \). The factorization process allows for a finer-grained representation of the graph, where different latent factors provide insights into various aspects of node interactions and relationships.
Other notations and the details of their usage will be clarified as they arise in subsequent discussions.


\subsection{Latent Factor Learning}

\label{subsec_3.2: latent_factor}
\Rebuttal{In this subsection, we present our probabilistic latent factor learning module, which is specifically designed to overcome the convergence challenges in generative disentangled representation learning. Unlike conventional approaches that directly apply existing factor learning methods, we reformulate the probabilistic graph generation process as a structural inductive bias to guide subsequent mask modeling.}
\Rebuttal{The specific approach involves modeling the distribution of nodes and edges, utilizing the generative process under Bernoulli-Poisson link , which is suitable for modelling the latent count variables \cite{zhou2015infinite, duan2021sawtooth}, can be described as}:

\begin{align}
    {\mbox{M}_{uv}} &\sim  \mbox{Poisson}( {\sum\nolimits_{k = 1}^K {{\gamma _k}{z_{uk}}{z_{vk}}} } ) \label{Eq_1: M_sample} \\   
    {z_{uk}} &\sim  \mbox{Gamma}\left( {\alpha, \beta} \right), u,v \in \left[ {1,N} \right] \label{Eq_2: z_gamma}
\end{align}

where ${K}$ is the predefined number of latent factors, and $u$ and $v$ are the indices of the nodes. Here, $\mbox{M}_{uv}$ is the latent count variable between node $u$ and $v$; $\gamma_k$ is a positive factor activation level indicator, which measures the node interaction frequency via factor $k$; 
$z_{uk}$ is a positive latent variable for node $u$, which measures how strongly node $u$ is affiliated with factor $k$. 
The prior distribution of latent factor variable $z_{uk}$ is set to Gamma distribution, where $\alpha$ and $\beta$ are normally set to 1.
\Rebuttal{The Gamma distribution is chosen not only for its non-negativity but also for its inherent sparsity-inducing properties \cite{wangnon2024, huenhancing}, which naturally frames the factorization as a non-negative matrix factorization problem, enhancing both identifiability and interpretability of the learned factors. We theoretically justify this benefit in Sec. \ref{subsubsec: identifiable_theory}.} 
Accordingly, the intuitive explanation for this generative process is that, with $z_{uk}$ and $z_{vk}$ measuring how strongly node $u$ and $v$ are affiliated with the $k$-th factor, respectively, the product ${\gamma _k}{z_{uk}}{z_{vk}}$ measures how strongly nodes $u$ and $v$ are connected due to their affiliations with the $k$-th factor. 


\paragraph{Node Factorization} \label{subsubsec_3.2.1: node_partition}
\Rebuttal{To obtain well-separated factor-specific subgraphs that can effectively guide the subsequent mask modeling, we derive factor-specific node sets through a carefully designed factorization process. }Equation \ref{Eq_1: M_sample} can be further augmented as follows:
\begin{equation}
\label{eq: 3.2.1_muv}
    {\mbox{M}_{uv}} = \sum\nolimits_k^K {{\mbox{M}_{ukv}}}, \, \  {\mbox{M}_{ukv}} \sim \mbox{Poisson}\left( {{\gamma _k}{z_{uk}}{z_{vk}}} \right)    
\end{equation}

where $\mbox{M}_{ukv}$ represents how often nodes $u$ and $v$ interact due to their affiliations with the $k$-th factor. 
To represent how often node $u$ is affiliated with the $k$-th factor, we further introduce the latent count
$ {\mbox{M}_{uk \cdot } = \sum\nolimits_{v \ne u} {{\mbox{M}_{ukv}}} }$.
Then, we can soft assign node $u$ to multiple factors in $\{ k:{\mbox{M}_{uk \cdot }}\}  \ge 1$, or hard assign node $u$ to a single factor using ${{\mathop {\arg \max }\limits_k \left( {{\mbox{M}_{uk \cdot }}} \right)}}$.
However, our experiments show that the soft assignment method results in significant overlap among node sets from different factor groups, diminishing the distinctiveness. 
Note that a previous study addressed a similar issue by selecting the top-k most attended regions \cite{kakogeorgiou2022hide}.
Thus, we choose the hard assign strategy to factorize the graph node set $V$ into factor-specific node sets $\{ V^{(1)}, V^{(2)},\cdots, V^{(K)} \}$.

\paragraph{Edge Factorization}
\label{subsubsec_3.2.2: edge_partition}

\Rebuttal{For edge factorization, we seek a \emph{factor-specific yet fully differentiable} construction that is compatible with end-to-end training. A straightforward but brittle choice is \emph{hard masking}, which keeps an edge only when both endpoints belong to the $k$-th node set $V^{(k)}$ (and removes it otherwise):}

\begin{equation} \label{eq: edge_1} A_{uv}^{\left( k \right)} = \left\{ \begin{array}{l} {A_{uv}},{\rm{ }} \: \forall \: u,v \in {V^{\left( k \right)}}; \: u,v \in \left[ {1 ,N} \right] \\ 0,{\rm{ }} \:\:\: \exists \: u,v \notin {V^{\left( k \right)}}; \: u,v \in \left[ {1 ,N} \right] \end{array} \right. \end{equation}

\Rebuttal{While Eq.~\eqref{eq: edge_1} enforces factor specificity, it is non-differentiable with respect to factor assignments and cannot recover from early partitioning errors. To retain trainability, we adopt a \emph{soft, positive-weighted} decomposition of the global adjacency $A$ \cite{he2022variational}:}

\begin{equation} \label{eq: edge_2} A_{uv}^{\left( k \right)} = {A_{uv}} \cdot \frac{{\exp \left( {{\gamma _k}{z_{uk}}{z_{vk}}} \right)}}{{\sum\nolimits_{k'}^{} {\exp \left( {{\gamma _{k'}}{z_{uk'}}{z_{vk'}}} \right)} }},\ \ \end{equation}

\Rebuttal{where $z_{uk}$ denotes node $u$’s affinity to factor $k$ and $\gamma_k$ is a learnable (or fixed) temperature. Applying Eq.~\eqref{eq: edge_2} to all node pairs yields $K$ weighted adjacency matrices ${A^{(k)}}_{k=1}^{K}$ of the same size as $A$, with $\sum_k A^{(k)}!=!A$. Compared with the hard mask in Eq.~\eqref{eq: edge_1}, Eq.~\eqref{eq: edge_2} preserves gradients with respect to ${z,\gamma}$, keeps the masking stochastic yet factor-consistent, and can be jointly optimized with the encoder–decoder. \emph{Therefore, we adopt Eq.~\eqref{eq: edge_2} for edge factorization.}}

\paragraph{Variational Inference}
\label{subsubsec_3.2.3: VI_epm}
\Rebuttal{The key to our framework's stability lies in the effective approximation of the posterior distribution of latent factors. We employ a Weibull variational encoder specifically designed to handle the non-negative constraints of our Gamma priors while ensuring stable gradient propagation during the joint training with the masked autoencoder.}
Denoting $z_u = (z_{u1}, ..., z_{uK}), z_u \in \mathbb{R}_{+}^{K}$, which measures how strongly node $u$ is affiliated with all the $K$ latent factors, we adopt a Weibull variational graph encoder \cite{zhang2018whai, he2022variational}:
\begin{align}
\label{eq: VI_1} 
    \begin{split} 
    q(z _{u} \mid A, X ) &= \mbox{Weibull}(k_u, \lambda _u), \\
    (k_u, \lambda _u) = \mbox{GNN}_{ \scalebox{0.7}{\mbox{EPM}}} &(A, X), \quad u \in \left[ {1,N} \right]
    \end{split}
\end{align}
where ${\mbox{GNN}_{ \scalebox{0.7}{\mbox{EPM}}}(\cdot)}$ stands for graph neural networks, and we select a two-layer Graph Convolution Network (\ie, GCN \cite{kipf2016semi}) for our models; ${k_u},{\lambda _u} \in \mathbb{R}_ + ^K$ are the shape and scale parameters of the variational Weibull distribution, respectively. The latent variable $ z _{u}$ can be conveniently reparameterized as:
\begin{equation}
\begin{split}
 z _{u} = \lambda _{u} {( - \ln (1 - \varepsilon ))^{1/k _{u}}},~~\varepsilon \sim \mbox{Uniform}(0,1)
\end{split}
\label{eq: VI_2}
\end{equation}
\Rebuttal{This variational formulation is not merely a technical choice but a crucial component that enables the seamless integration of probabilistic factor learning with generative mask modeling. The optimization objective of latent factor learning phase can be achieved by maximizing the evidence lower bound (ELBO) of the log marginal likelihood of edge $\log p(A)$, which serves as a bridge between factor discovery and graph reconstruction, allowing both components to be optimized jointly rather than as separate modules. It can be computed as:}
\begin{equation} \label{eq: VI_3}
\begin{split}
\mathcal{L_{\text{z}}} & =  \mathbb{E}_{q( Z  \,|\, A, X )} \left[ \ln  p\left(A \,|\, Z \right) \right] \\
&-\sum_{u = 1}^N  {{\mathbb{E}_{q( \boldsymbol{z}_u \,|\, A, X )}}\left[ \ln \frac{{q( \boldsymbol{z}_u \,|\, A, X )}}{p(\boldsymbol{z}_u)} \right]} 
\end{split}
\end{equation}
where the first term is the expected log-likelihood or reconstruction error of edge, and the second term is the Kullback–Leibler (KL) divergence that constrains $q(z_u)$ to be close to its prior $p(z_u)$.
The analytical expression for the KL divergence and the straightforward reparameterization of the Weibull distribution simplify the gradient estimation of the ELBO concerning the decoder parameters and other parameters in the inference network.

\subsection{Disentangled Graph Masked Autoencoder}
\label{subsec_3.3: disentangled GMAE}
With the latent factor learning phase discussed in \ref{subsec_3.2: latent_factor}, the graph can be factorized into a series of factor-specific subgraphs ${\{G_1, G_2,..., G_K\}}$ via the latent factor $z$.
To incorporate the disentangled information encapsulated in $z$ into the graph masked autoencoder, we propose the Disentangled Graph Masked Autoencoder in this section.
Specifically, this section will first introduce the latent factor-wise GMAE and the graph-level GMAE.
Following that, we provide details on the comprehensive training algorithm, along with the associated inference process.

\subsubsection{Latent Factor-wise Graph Masked Autoencoder}
\label{subsubsec_3.3.1: DiG MGAE}
To capture disentangled patterns within the latent factor $z$
, for each latent subgraph ${{ \mathcal{G}_k =(V^{(k)},A^{(k)}, X^{(k)})}}$, 
the latent factor-wise GMAE can be described as:
\begin{equation}
\label{eq: 3.3_com_1}
    {H_{d}^{(k)}} = \mbox{GNN}_{\scalebox{0.8}{\mbox{enc}}}( {{A^{(k)}},{{\bar X}^{(k)}}} ), \ \ 
    {{\tilde X}^{d}} =\mbox{GNN}_{\scalebox{0.8}{\mbox{dec}}}( A, H_{d} ).
\end{equation}
where ${{\bar X}^{(k)}}$ is the masked node feature matrix for the $k$-th latent factor, and ${{\tilde X}^{d}}$ denotes the reconstructed node features. $\scalebox{0.95}{\mbox{GNN}}_{\scalebox{0.8}{\mbox{enc}}}(.)$ and   $\scalebox{0.95}{\mbox{GNN}}_{\scalebox{0.8}{\mbox{dec}}}(.)$ are the graph encoder and  decoder, respectively;
${{H_{d}^{(k)}\in\mathbb{R}^{N \times D}}}$ are factor-wise hidden representations, and $H_{d} = H_{d}^{(1)}  \oplus  H_{d}^{(2)} \cdots \oplus H_{d}^{(K)}$. After the concatenation operation $ \oplus $ in feature dimension, the multi factor-wise hidden representation becomes
$H_{d} \in \mathbb{R}^{N \times {(K \cdot D)}}$, which is used as the input of ${\mbox{GNN}}_{\scalebox{0.8}{\mbox{dec}}}(.)$.

Regarding the mask operation, we uniformly random sample a subset of nodes ${{\bar V}^{(k)} \in V^{(k)}}$ and mask each of their features with a mask token, such as a learnable vector ${X_{[M]} \in \mathbb{R}^{d}}$. Thus, the node feature in the masked feature matrix can be defined as:
\begin{equation}
\label{eq: com_2}
\begin{split}
{\bar X}_{i}^{\left( k \right)} = \left\{ \begin{array}{l}
{X_{[M]}};\ \  {\rm{  }} v_i \in {\bar V^{\left( k \right)}}\:\:; \\ 
X_i \:\:\:\:; \ \ v_i \notin {\bar V^{\left( k \right)}}.
\end{array} \right.
\end{split}
\end{equation}
The objective of latent factor-wise GMAE is to reconstruct the masked features of nodes in ${\bar V}^{(k)}$ given the partially observed node signals ${\bar X}^{(k)}$ and the input adjacency matrix $A^{(k)}$. 
Another crucial component of the GMAE is the feature reconstruction criterion, often used in language as cross-entropy error \cite{devlin2018bert} and in the image as mean squared error  \cite{he2022masked}. 
However, texts and images typically involve tokenized input features, whereas graph autoencoders (GAE) do not have a universal tokenizer.
We adopt the scored cosine error of GraphMAE \cite{hou2022graphmae} as the loss function. 
Generally, given the original feature ${X^{(k)}}$ and reconstructed node feature ${ {\tilde X^{(k)}}}$, the defined SCE is:
\begin{equation}
\label{Eq_SCE}
\begin{split}
{\mathcal{L}_{\scalebox{0.7}{D}}} = \frac{1}{{| {{{\bar V}}}|}}{\sum\limits_{i \in {{\bar V}}} {\left( 1 - \frac{{X_{i}^{T} \tilde X^{d}_{i}}}{{\| {X_{i}} \| \cdot \| {\tilde X^{d}_{i}} \|}} \right)^\gamma}  },~~   \gamma  \ge 1    
\end{split}
\end{equation}
where $\bar V = {{\bar V}^{(1)}} \cup {{\bar V}^{(2)}} ... \cup {{\bar V}^{(K)}}$ and Equation~\ref{Eq_SCE} is averaged over all masked nodes.The scaling factor $\gamma $ is a hyper-parameter adjustable over different datasets. 
This scaling technique could also be viewed as adaptive sample reweighting, and the weight of each sample is adjusted with the reconstruction
error. This error is also famous in the field of supervised object detection as the focal loss~\cite{lin2017focal}.

\subsubsection{Graph-level Graph Mask Autoencoder}
\label{subsubsec_3.3.2: graph_level MGAE}
For the node classification task, we have integrated graph-level GMAE into DiGGR. We provide a detailed experimental analysis and explanation for this difference in Section \ref{appendix_a.1.2: ablation_repre}.
The graph-level masked graph autoencoder is designed with the aim of further capturing the global patterns, which can be designed as:
\begin{equation}
\label{eq: 3.3_graph_1}
    {H_{g}} = \mbox{GNN}_{\scalebox{0.8}{\mbox{enc}}}( {{A},{\bar X}} ), \:\:
    {{\tilde X}^{g}} = \mbox{GNN}_{\scalebox{0.8}{\mbox{dec}}}( A, H_{g} ).
\end{equation}

${\bar X}$ is the masked node feature matrix, whose mask can be generated by uniformly random sampling a subset of nodes $\tilde V \in V$, or obtained by concatenating the masks of all factor-specific groups $\tilde V = {{\bar V}^{(1)}} \cup {{\bar V}^{(2)}} ... \cup {{\bar V}^{(K)}}$.
The global hidden representation encoded by $\mbox{GNN}_{\scalebox{0.8}{\mbox{enc}}}(.)$ is $H_{g}$, 
which is then passed to the decoder.
Similar to Equation \ref{Eq_SCE}, we can define the graph-level reconstruction loss as:
\begin{equation}
\begin{split}
{\mathcal{L}_{\scalebox{0.7}{G}}} = \frac{1}{{| {{{\tilde V}}} |}}{\sum\limits_{i \in {{\tilde V}}} {( {1 - \frac{{X_i^{T}\tilde X_i^{g}}}{{\left\| {X_i} \right\| \cdot \| {\tilde X_i^{g}} \|}}} )^\gamma} },\ \ \gamma  \ge 1  .
\end{split}
\end{equation}
which is averaged over all masked nodes.


\subsection{Joint Training and Inference}
\label{subsec_3.4: Train&Infer}
Benefiting from the effective variational inference method, the proposed latent factor learning and disentangled graph masked autoencoder can be jointly trained in one framework.
We combine the aforementioned losses with three mixing coefficients $\lambda_d$, $\lambda_g$ and $\lambda_{z}$ during training, and the loss for joint training can be written as
\begin{equation}
\label{Eq_loss}
\begin{split}
\mathcal{L} = 
\lambda_d \cdot {\mathcal{L}_{\scalebox{0.7}{D}}}  + 
\lambda_g \cdot 
{\mathcal{L}_{\scalebox{0.7}{G}}} + 
\lambda_{z} \cdot
{\mathcal{L}_{\scalebox{0.7}{z}}}.
\end{split}
\end{equation}
Since Weibull distributions have easy reparameterization functions, these parameters can be jointly trained by stochastic gradient descent with low-variance gradient estimation. 
We summarize the training algorithm at Algorithm~\ref{alg_DiGGR}.
For downstream applications, the encoder is applied to the input graph without any masking in the inference stage. 
The generated factor-wise node embeddings $H_d$ and graph-level embeddings $H_g$ can either be concatenated in the feature dimensions or used separately. The resulting final representation $H$ can be employed for various graph learning tasks, such as node classification and graph classification. For graph-level tasks, we use a non-parameterized graph pooling (readout) function, \eg,  MaxPooling and MeanPooling to obtain the graph-level representation.

\begin{algorithm}[t]
  \caption{The Overall Training Algorithm of DiGGR}
  \label{alg_DiGGR}
\begin{algorithmic}[1]
  \STATE \textbf{Input}: Graph $\mathcal{G} = \{V, A, X \}$; latent factor number $K$.
  \STATE \textbf{Parameters}: 
  $\Theta$ in the inference network of Latent Factor Learning phase, 
  $\mathbf{\Omega}$ in the encoding network of DiGGR,
  $\mathbf{\Psi}$ in the decoding network of DiGGR. 
  \STATE Initialize $\Theta$, $\mathbf{\Omega}$, and $\mathbf{\Psi}$;
  \FOR{$\text{iter = 1,2,} \cdot \cdot \cdot $} 
    \STATE Infer the \textit{variational posterior} of $\mathbf{z_u}$ based on Eq.~\ref{eq: VI_1}; 
    \STATE Sample latent factors $\mathbf{z_u}$ from the variational posterior according to Eq.~\ref{eq: VI_2};
    \STATE Factorize the graph $\mathcal{G}$ into $K$ factor-wise groups $\mathbf{\{\mathcal{G}^{(k)}\}_{k=1}^{K}}$ by node and edge factorization methods;
    \STATE Encode $\mathbf{\{\mathcal{G}^{(k)}\}_{k=1}^{K}}$ via latent factor-wise Graph Masked Autoencoder according to Eq.~\ref{eq: 3.3_com_1};
    \STATE Encode $\mathcal{G}$ via graph-level Graph Masked Autoencoder according to Eq.~\ref{eq: 3.3_graph_1};
    \STATE Calculate 
    $\nabla_{\Theta, \mathbf{\Omega}, \mathbf{\Psi}}\mathcal{L}({\Theta, \mathbf{\Omega}, \mathbf{\Psi}}; \mathcal{G}) $
    according to Eq.~\ref{Eq_loss}, and update parameters $\Theta$, $\mathbf{\Omega}$, and $\mathbf{\Psi}$ jointly.
  \ENDFOR
\end{algorithmic}
\end{algorithm}

\textbf{Time and space complexity:}
\Rebuttalmr{Let $N$, $M$, and $K$ denote the numbers of nodes, edges, and latent factors, and let $F$ be the feature dimension. Let $L_1$–$L_4$ be the layer numbers of the latent-factor encoder, factor-wise GMAE encoder, graph-level GMAE encoder, and decoder, respectively. }
With the factor-wise encoder hidden size constrained to $1/K$ of the baseline, the time complexity for training DiGGR can be expressed as $O((L_1 + L_2 + L_3)MF + (L_1 + L_2/K + L_3)NF^2 + N^2F + L_4NF^2)$, and the space complexity is $O((L_1 +L_2 +L_3 + L_4)NF +KM +(L1 +L2/K +L3 + L_4)F^2)$, with $O((L_1 + L_2/K + L_3 + L_4)F^2)$ attributed to model parameters. 
We utilize the Bayesian factor model in our approach to reconstruct edges. Its time complexity aligns with that of variational inference in \cite{li2023seegera}, predominantly at $O(N^2F)$. Thus, the complexity of DiGGR is comparable to previous works. \Rebuttal{The computation experiment is provided in Table \ref{tab_9: computation_analysis}.}

\subsection{Theoretical Analysis}

\subsubsection{\Rebuttal{Identifiability of the learned latent factor}}\label{subsubsec: identifiable_theory}
\Rebuttal{We provide theoretical guarantees for DiGGR from the perspective of identifiable features. In latent-variable models, identifiability enhances the ability to obtain disentangled representations and is a prerequisite for achieving disentanglement \cite{khemakhem2020variational}, we would further justify the choice of Gamma prior under this perspective.
The generative process of latent factor in Eq. \ref{Eq_1: M_sample} can be viewed as non-negative matrix factorization (\textbf{NMF}) \cite{wang2012nonnegative, huenhancing}. In detail, Eq. \ref{eq: VI_1} ensures that the expectation of Weibull distribution is proportional to its scale parameter. Therefore, instead of sampling $z_{uk}$ and $z_{vk}$ from their respective distributions, substituting their expectations makes the mapping equivalent to that of standard NMF, resulting in $M_{uv} = \sum_{k=1}^{K} z_{uk}z_{vk}=z_uz_v $ and the following objective:}

\Rebuttal{\begin{equation}
    \label{eq_15: NMF}
        {\mathop{\min}_{z_{u} \geq 0, z_{v} \geq 0} {\lVert M_{uv} - z_{u}z_{v} \rVert}_F^2}
\end{equation}}

\Rebuttal{Many studies in the NMF literature aim to establish, in practical applications, the uniqueness and identifiability of the decomposition ($z_u z_v$) up to trivial ambiguities (e.g., permutation and scaling). However, the conditions for full identifiability are often overly stringent \cite{locatello2019challenging}. \cite{gillis2023partial} shows that ($z_u z_v$) admits partial identifiability provided the following two conditions hold:
\textbf{\textit{i) Selective Window:}} There exists a row of $z_v$, say the $j$-th, such that $z_v(j, :)=\alpha e_{(k)}^T$ for $\alpha > 0$, where $e_{(k)}^T$ represents the $k$-th standard row vector in vector space. \textbf{\textit{ii) Sparsity Constraint}}: The $k$-th column of $z_v$ contains at least $r-1$ entries equal to zero, where $r$ is the rank of $M$. We formally state the assumptions below:}
\Rebuttal{\begin{assumption}
    \label{assump_1}
         $z_v$ and $z_u$ in Eq. \ref{eq_15: NMF} satisfy the \textbf{\textit{Selective Window}} condition and the \textbf{\textit{Sparsity Constraint}}. 
\end{assumption}}
\Rebuttal{We now show how the latent factor of DiGGR fits the above assumption.\textbf{\textit{ i) For the Selective Window}}: In brief, satisfying this assumption is equivalent to requiring that there exists a node (u) in the dataset that is associated with exactly one latent factor. This assumption is reasonable in many practical settings; for example, topic modeling often posits the existence of anchor words that are tied to a single topic \cite{arora2013practical}, while \cite{wangnon2024} assumes that every latent class corresponds to a unique sample. By contrast, we require only the existence of a unique sample for some latent factor, which is more readily attainable in practice. \textbf{\textit{ii) For the Sparsity Constraint}}: This assumption emphasizes the sparsity of $z_v$, which also \textbf{motivates the use of a Gamma prior} in Eq.~\eqref{Eq_2: z_gamma}. The Gamma prior is inherently \textbf{nonnegative}; moreover, by setting its shape parameter to 1, its probability mass concentrates near zero, thereby enforcing \textbf{sparser representations} during training \cite{zhou2015infinite}. Under this assumption, DiGGR promotes the partial identifiability of the learned latent factors and thus enhances their disentanglement capability. Our experiments in Table~\ref{tab_5: sepin@cora} corroborate this finding, with the representations learned under the Gamma prior being more identifiable and disentangled.}


\subsubsection{Effectiveness of Latent factor guided} \label{subsubsec: thoery_gae}
DiGGR is built on a graph autoencoder (GAE)-based framework. Recent studies \cite{zhang2022mask, li2023s} have demonstrated a direct connection between GAE and contrastive learning through a specific form of the objective function. The loss function can be rewritten as follows: 

\begin{equation}
	\centering
	\label{eq_theory:gae}
	 L^+ = \frac{1}{|\varepsilon^+|}\sum_{(u, v) \in V^+}{\log f_{dec}(h_u, h_v)} $$ $$ L^- = \frac{1}{|\varepsilon^-|}\sum_{(u', v') \in V^-}{\log (1 - f_{dec}(h_{u'}, h_{v'}))} $$ $$ L_{GAE} = -(L^+ + L^-) 
\end{equation}

where $h_u$ and $h_v$ are the node representations of node $u$ and node $v$ obtained from an encoder $f_{enc}$ respectively (eg., a GNN); $\varepsilon^+$ is a set of positive edges while $\varepsilon^-$ is a set of negative edges sampled from graph, and $f_{dec}$ is a decoder; Typically, $\varepsilon^+ = \varepsilon$.
Building on recent advances in information-theoretic approaches to contrastive learning \cite{tsai2020self, tian2020makes}, a recent study \cite{li2023s} suggests that for SSL pretraining to succeed in downstream tasks, task-irrelevant information must be reasonably controlled. Therefore, the following proposition is put forward:

\begin{proposition}[\cite{li2023s}]
    \label{props_1}
    The task irrelevant information $I(U; V| T)$ of GAE can be lower bounded with: 
    $$ I(U; V| T) \geq \frac{(E[N_{uv}^{k}])^2}{N_k} . \gamma^2 $$ 
    Minimizing the aforementioned $L_{GAE}$ is in population equivalent to maximizing the mutual information between the k-hop subgraphs of adjacent nodes, and the redundancy of GAE scales almost linearly with the size of overlapping subgraphs.
\end{proposition}
The above proposition has been proved in detail in \cite{li2023s}, where $I(.;.)$ is the mutual information, $U$ and $V$ be random variables of the two contrasting views, and $T$ denote the target of the downstream task. $N_{uv}^{k}$ is the size of the overlapping subgraph of $G^k(u)$ and $G^k(v)$, and the expectation is taken with respect to the generating distribution of the graph and the randomness in choosing $u$ and $v$.
According to this lower bound, \Rebuttal{we need to reduce the task-irrelevant redundancy to design a better graph SSL methods}. In DiGGR, we first factorize the input graph based on latent factor learning before feeding it into the masked autoencoder.
\begin{figure}[tbp]
    \centering
    \includegraphics[width=1.0\linewidth]{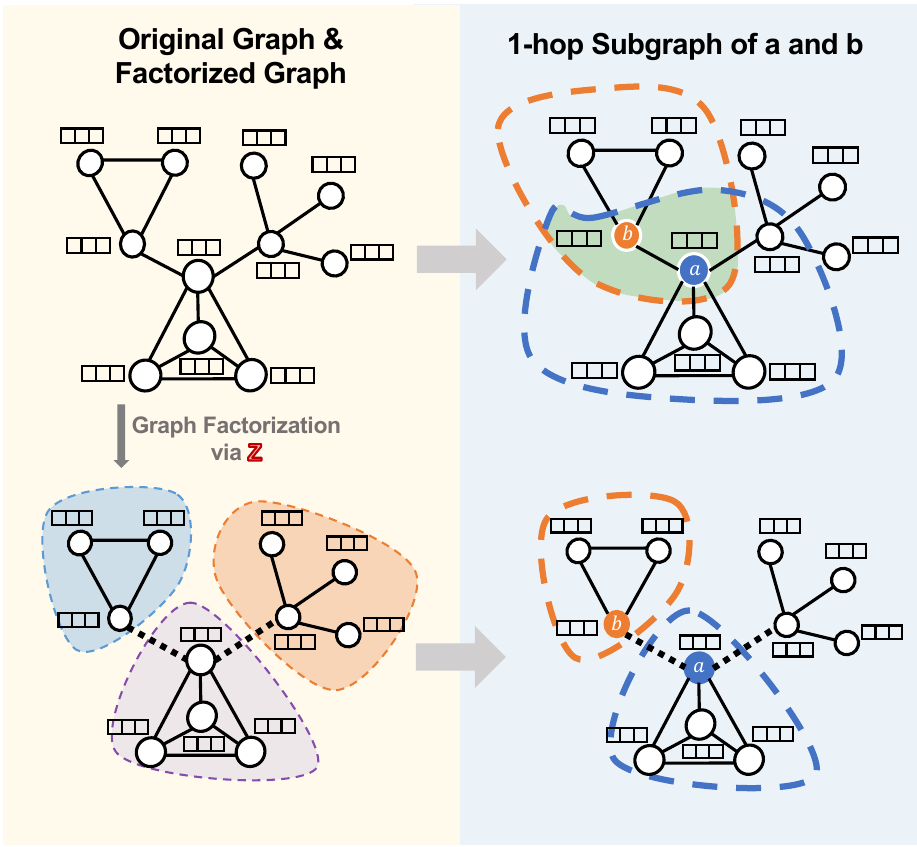}
    \caption{
    	\Rebuttalmr{Graph factorization reduces 1-hop subgraph overlap. 
    	In the original graph, the 1-hop neighborhoods of nodes $a$ and $b$ overlap (green region); after factorization, their connection is removed and the overlap is reduced.}}
    \label{fig_appendix: theoretical}
\end{figure}
Take Fig. \ref{fig_appendix: theoretical} as an example. Nodes $a$ and $b$ have overlapping 1-hop subgraphs. However, after graph factorization, the connection between $a$ and $b$ is severed, thereby reducing large-scale subgraph overlap and lowering the lower bound of task-irrelevant information. 
We also conduct the corresponding quantitative analysis in later experiment Sec.\ref{F.a: disentangle_anylsis}. As shown in Table~\ref{table_4.3: disentangled}, after factorization, the latent factor groups extracted by DiGGR exhibit lower normalized mutual information (NMI) \Rebuttal{while exhibiting performance gains}, indicating reduced overlap between the latent factor groups.This result aligns with our theoretical analysis and highlights the advantages of our proposed method.

\section{Experiments}
\label{sec_4: exp}
We compare the proposed self-supervised framework DiGGR against related baselines on two fundamental tasks: unsupervised representation learning on \textit{node classification} and \textit{graph classification}.
We evaluate DiGGR on 15 benchmarks. For node classification, we use 3 citation networks (Cora, Citeseer, Pubmed \cite{yang2016revisiting}), and protein-protein interaction networks (PPI) \cite{hamilton2017inductive}. 
\Rebuttal{For node classification, we also evaluated the results on four heterophilous datasets (Texas, Wisconsin, Cornell, Actor \cite{pei2020geom}).}
For graph classification, we use 3 bioinformatics datasets (MUTAG, NCI1, PROTEINS) and 4 social network datasets (IMDB-BINARY, IMDB-MULTI, REDDIT-BINARY and COLLAB)\cite{yanardag2015deep}. 

\subsection{Implementation Details}
\label{appendix_a.2: exp details}

\Rebuttalmr{The node features for citation networks (Cora, Citeseer, Pubmed) are standard bag-of-words document representations, while PPI adopts the 50-dimensional gene-set features released in prior work. For heterophilous graphs (Texas, Wisconsin, Cornell, Actor), we follow the official train/validation/test splits used in \cite{li2022finding}. In graph classification, MUTAG, PROTEINS and NCI1 use one-hot node labels as features, whereas IMDB-B/M, REDDIT-B and COLLAB use one-hot degree features capped at 400 following GraphMAE \cite{hou2022graphmae}.}

\begin{table*}[h]
	\centering
	\caption{Experiment results in unsupervised representation learning for graph classification. We report accuracy (\%) for all datasets. The optimal outcomes for methods, excluding supervised approaches (GIN and DiffPool), on each dataset are emphasized in \textbf{bold}.}
	\label{table: Graph Classification}
	\renewcommand{\arraystretch}{1.1} 
	\scalebox{1.0}{
		\begin{tabular}{ p{2.1cm}<{\centering} p{1.8cm}<{\centering}  p{1.8cm}<{\centering} p{1.8cm}<{\centering} p{1.8cm}<{\centering}
				p{1.8cm}<{\centering} p{1.8cm}<{\centering} p{1.8cm}<{\centering}}
			\toprule
			\textbf{Methods} & \textbf{IMDB-B} & \textbf{IMDB-M} & \textbf{MUTAG} & \textbf{NCI1} & \textbf{REDDIT-B} & \textbf{PROTEINS} & \textbf{COLLAB} \\
			\midrule
			GIN \cite{xu2018powerful}& 75.1 $\pm$ 5.1 & 52.3 $\pm$ 2.8 & 89.4 $\pm$ 5.6 & 82.7 $\pm$ 1.7 & 92.4 $\pm$ 2.5 & 76.2 $\pm$ 2.8 & 80.2 $\pm$ 1.9 \\
			DiffPool \cite{ying2018hierarchical}& 72.6 $\pm$ 3.9 & - & 85.0 $\pm$ 10.3 & - & 92.1 $\pm$ 2.6 & 75.1 $\pm$ 3.5 & 78.9 $\pm$ 2.3 \\
			VEPM \cite{he2022variational}& 76.7 $\pm$ 3.1 & 54.1 $\pm$ 2.1 & 93.6 $\pm$ 3.4 & 83.9 $\pm$ 1.8 & 90.5 $\pm$ 1.8 & 80.5 $\pm$ 2.8 & - \\
			\Rebuttal{DisenSemi \cite{wang2024disensemi}} & \Rebuttal{76.7 $\pm$ 0.7} & \Rebuttal{52.5 $\pm$ 0.5} & \Rebuttal{92.6 $\pm$ 0.6} & \Rebuttal{78.9 $\pm$ 0.4} & \Rebuttal{93.2 $\pm$ 0.3} & \Rebuttal{78.4 $\pm$ 0.5} & \Rebuttal{81.5 $\pm$ 0.3} \\
			\midrule
			WL \cite{shervashidze2011weisfeiler} & 72.30 $\pm$ 3.44 & 46.95 $\pm$ 0.46 & 80.72 $\pm$ 3.00 & 80.31 $\pm$ 0.46 & 68.82 $\pm$ 0.41 & 72.92 $\pm$ 0.56 & - \\
			DGK \cite{yanardag2015deep} & 66.96 $\pm$ 0.56 & 44.55 $\pm$ 0.52 & 87.44 $\pm$ 2.72 & 80.31 $\pm$ 0.46 & 78.04 $\pm$ 0.39 & 73.30 $\pm$ 0.82 & 73.09 $\pm$ 0.25 \\
			\midrule
			Infograph \cite{sun2019infograph} & 73.03 $\pm$ 0.87 & 49.69 $\pm$ 0.53 & 89.01 $\pm$ 1.13 & 76.20 $\pm$ 1.06 & 82.50 $\pm$ 1.42 & 74.44 $\pm$ 0.31 & 70.65 $\pm$ 1.13 \\
			GraphCL \cite{you2020graph} & 71.14 $\pm$ 0.44 & 48.58 $\pm$ 0.67 & 86.80 $\pm$ 1.34 & 77.87 $\pm$ 0.41 & \textbf{89.53} $\pm$ 0.84 & 74.39 $\pm$ 0.45 & 71.36 $\pm$ 1.15 \\
			JOAO \cite{you2021graph} & 70.21 $\pm$ 3.08 & 49.20 $\pm$ 0.77 & 87.35 $\pm$ 1.02 & 78.07 $\pm$ 0.47 & 85.29 $\pm$ 1.35 & 74.55 $\pm$ 0.41 & 69.50 $\pm$ 0.36 \\
			GCC \cite{qiu2020gcc} & 72.0 & 49.4 & - & - & 89.9 & - & 78.9 \\
			MVGRL \cite{hassani2020contrastive} & 74.20 $\pm$ 0.70 & 51.20 $\pm$ 0.50 & 89.70 $\pm$ 1.10 & - & 84.50 $\pm$ 0.60 & - & - \\
			InfoGCL \cite{xu2021infogcl} & 75.10 $\pm$ 0.90 & 51.40 $\pm$ 0.80 & 91.20 $\pm$ 1.30 & 80.20 $\pm$ 0.60 & - & - & 80.00 $\pm$ 1.30 \\
			\Rebuttal{DGCL \cite{li2021disentangled}} & \Rebuttal{75.9 $\pm$ 0.7}  & \Rebuttal{51.9 $\pm$ 0.4} & \Rebuttal{92.1 $\pm$ 0.8} &  \Rebuttal{81.9 $\pm$ 0.2} & \Rebuttal{91.8 $\pm$ 0.2}  & \Rebuttal{76.4 $\pm$ 0.5} & \Rebuttal{81.2 $\pm$ 0.3}\\
			\Rebuttal{AutoGCL \cite{yin2022autogcl}} & \Rebuttal{73.3 $\pm$ 0.40}  & \Rebuttal{-} & \Rebuttal{88.64 $\pm$ 1.08} &  \Rebuttal{82.00 $\pm$ 0.29} & \Rebuttal{88.58 $\pm$ 1.49}  & \Rebuttal{75.8 $\pm$ 0.36} & \Rebuttal{70.12 $\pm$ 0.68} \\
			\midrule
			graph2vec \cite{narayanan2017graph2vec} & 71.10 $\pm$ 0.54 & 50.44 $\pm$ 0.87 & 83.15 $\pm$ 9.25 & 73.22 $\pm$ 1.81 & 75.78 $\pm$ 1.03 & 73.30 $\pm$ 2.05 & - \\
			GAE \cite{kipf2016variational} & 52.1 $\pm$ 0.2 & - & 84.0 $\pm$ 0.6 & 73.3 $\pm$ 0.6 & 74.8 $\pm$ 0.2 & 74.1 $\pm$ 0.5 & - \\
			VGAE \cite{kipf2016variational} & 52.1 $\pm$ 0.2 & - & 84.4 $\pm$ 0.6 & 73.7 $\pm$ 0.3 & 74.8 $\pm$ 0.2 & 74.8 $\pm$ 0.2 & - \\
			GraphMAE \cite{hou2022graphmae} & 75.52 $\pm$ 0.66 & 51.63 $\pm$ 0.52 & 88.19 $\pm$ 1.26 & 80.40 $\pm$ 0.30 & 88.01 $\pm$ 0.19 & 75.30 $\pm$ 0.39 & 80.32 $\pm$ 0.46 \\
			GraphMAE2 \cite{hou2023graphmae2}& 73.88 $\pm$ 0.53 & 51.80 $\pm$ 0.60 & 86.63 $\pm$ 1.33 & 78.56 $\pm$ 0.26 & 76.84 $\pm$ 0.21 & 74.86 $\pm$ 0.34 & 77.59 $\pm$ 0.22 \\
			\midrule
			\rowcolor{gray!30}
			\textbf{DiGGR} & \textbf{77.68} $\pm$ 0.48 & \textbf{54.69} $\pm$ 2.06 & 88.72 $\pm$ 1.03 & \textbf{81.23} $\pm$ 0.40 & 88.19 $\pm$ 0.28 & \textbf{77.61} $\pm$ 0.12 & \textbf{83.76} $\pm$ 3.70 \\
			\bottomrule 
		\end{tabular} 
	}
\end{table*}

\subsection{Node Classification}
\label{subsec_4.1: node clss}
\begin{figure}[tbp]
	\centering
	\includegraphics[width=1.0\linewidth]{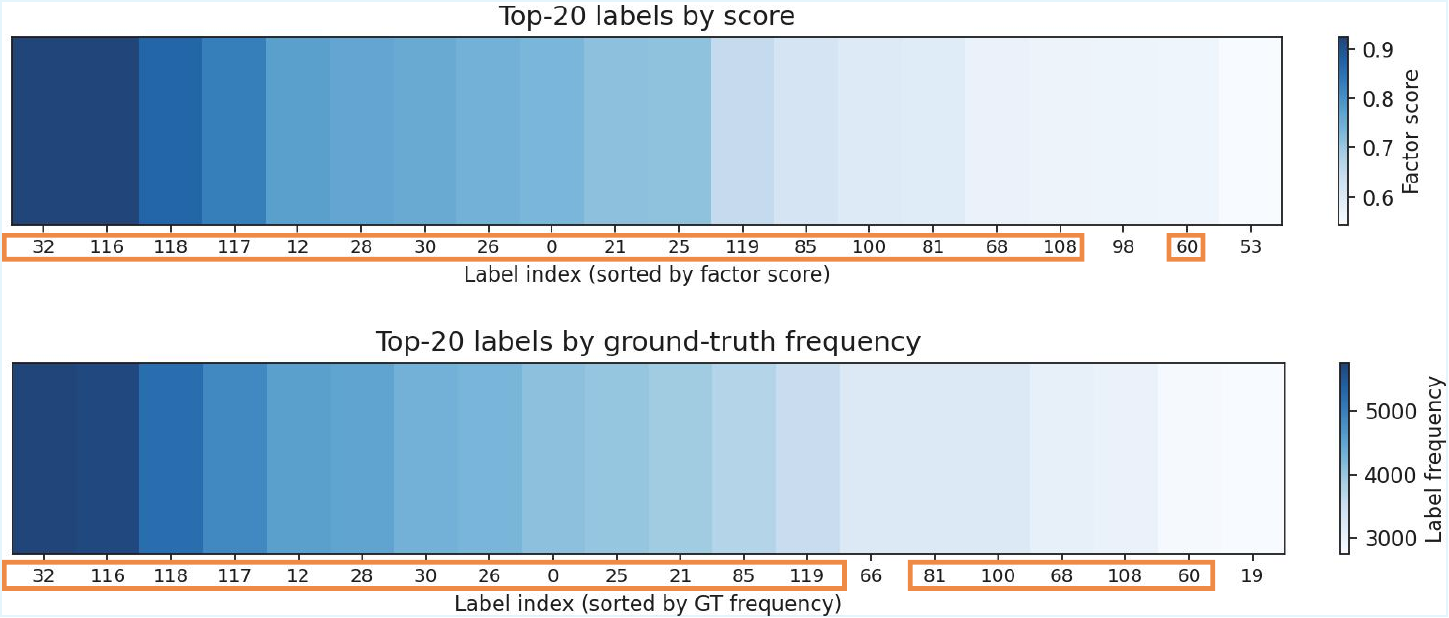}
	\caption{\Rebuttal{Correlation scores between latent factors $z$ and labels (top row; darker colors indicate stronger factor–label association) and the heatmap of the most frequent ground-truth labels (bottom row; darker colors indicate higher prevalence). The x-axis shows the corresponding label indices.}}
	\label{fig:ppi-factor-gt}
\end{figure}

The node classification task involves predicting the unknown node labels in networks. Cora, Citeseer, and Pubmed are employed for transductive learning, whereas PPI follows the inductive setup outlined in GraphSage \cite{hamilton2017inductive}.
For evaluation, we use the concatenated representations of $H_d$ and $H_g$ in the feature dimension for the downstream task. We then train a linear classifier and report the mean accuracy on the test nodes through 5 random initializations. The graph encoder $\mbox{GNN}_{\scalebox{0.8}{\mbox{enc}}}(.)$ and decoder $\mbox{GNN}_{\scalebox{0.8}{\mbox{dec}}}(.)$ are both specified as standard GAT \cite{velickovic2017graph}. We train the model using Adam Optimizer, and we use the cosine learning rate decay without warmup. We follow the public data splits of Cora, Citeseer, and Pubmed.
\Rebuttalmr{The baseline models for node classification can be divided into three categories:
$i)$ supervised methods, including GCN \cite{kipf2016semi}, GAT \cite{velickovic2017graph}, etc.;
$ii)$ contrastive learning methods, including MVGRL \cite{hassani2020contrastive}, GRACE \cite{zhu2020deep}, etc.;
$iii)$ generative learning methods, including GraphMAE \cite{hou2022graphmae}, VGAE \cite{kipf2016variational}, etc.}
Table~\ref{table: node_classification} summarizes node classification performance. \Rebuttal{For the comparison with \cite{zhang2023spectral}, we report the $\mathrm{SFA}_\mathrm{BT}$ variant, because DiGGR is a generative autoencoding pre-training method that does not rely on negative samples or large-batch statistics; $\mathrm{SFA}_\mathrm{BT}$ likewise omits negatives. This choice ensures methodological parity and a fair comparison.} DiGGR matches or surpasses supervised baselines on Cora, Citeseer and PPI, and remains competitive on Pubmed, indicating that disentangled generative pretraining can reach task-specific supervised models. 
On the multi-label PPI dataset, we further compare the 20 most frequent GO labels in the test set with the AUPRC of each learned factor (Fig. \ref{fig:ppi-factor-gt}). The high-response regions in the two heatmaps largely coincide, showing that dominant latent factors capture biologically meaningful protein functions.


\begin{table}[t]
	\centering
	\renewcommand{\arraystretch}{1.25}
	\caption{Node classification accuracy on limited-structure graph, where the best results are highlighted in \textbf{bold}. DiGGR has shown the comparable performance against others which specially designed for heterophilous benchmarks, demonstrating its generalization capability.}
	\label{tab: node_heter_clf}
	\scalebox{0.9}{
		\begin{tabular}{lcccc}
			\toprule
			\textbf{Methods} & \textbf{Texas}        & \textbf{Wisconsin}    & \textbf{Cornell} & \textbf{Actor} \\ \hline
			GraphMAE \cite{hou2022graphmae}        & 74.32 ± 4.56  & 74.71 ± 8.83   & 75.14 ± 9.65  & 28.58 ± 1.98    \\ \hline
			GREET \cite{liu2023beyond}        & 87.03 ± 2.36   & 84.90 ± 4.48   & 85.14 ± 4.87  & 36.55 ± 1.01   \\ 
			MUSE \cite{yuan2023muse} & 89.73 ± 2.79 & \textbf{88.24 ± 3.20} & 82.16 ± 3.42 & 38.55 ± 1.34 \\ 
			OGNN \cite{songordered2023}            & 86.22 ± 4.12          & 88.04 ± 3.63    & 87.03 ± 4.73 & 37.99 ± 1.00  \\
			LINKX \cite{yan2022two}           & 74.60 ± 8.37          & 75.49 ± 5.72     &  77.84 ± 5.81  & 36.10 ± 1.55 \\
			LRGNN \cite{liang2023predicting}           & \textbf{90.27 ± 4.49}           &88.23 ± 3.54    &   86.48 ± 5.65  & 37.34 ± 1.78 \\ \hline
			GraphACL \cite{xiao2023simple} & 71.08 ± 0.34 & 69.22 ± 0.40 & 59.33 ± 1.48 & 30.03 ± 0.13 \\ 
			DSSL  \cite{zhu2021dssl} & 62.11 ± 1.53 & - & 53.15 ± 1.28 & - \\ \hline
			\rowcolor{gray!30}
			\textbf{DiGGR}  & 85.95 ± 6.71 & 87.25 ± 4.41 &  \textbf{88.38 ± 6.84} & \textbf{45.35 ± 3.53} \\ \bottomrule
		\end{tabular}
	}
\end{table}

\begin{table}[t]
	\centering
	\caption{Experiment results for node classification. Micro-F1 score is reported for PPI, and accuracy for other datasets. The \textbf{best unsupervised} method scores in each dataset are highlighted in bold. \Rebuttal{We follow the common practice of recent graph representation learning papers and report the mean $\pm$ std under multiple random seeds to reflect the stability and statistical reliability of DiGGR. When the literature only provides a single result, we retain the value but do not directly compare it with the mean.}}
	\label{table: node_classification}
	\renewcommand{\arraystretch}{1.1} 
	\scalebox{0.85}{
		\begin{tabular}{cccccc}
			\toprule
			\textbf{Methods} & \textbf{Cora} & \textbf{Citeseer} & \textbf{Pubmed} & \textbf{PPI} \\
			\midrule
			GCN \cite{kipf2016semi} & 81.50 & 70.30 & 79.00 & 75.70 $\pm$ 0.10 \\
			GAT \cite{velickovic2017graph}  & 83.00 $\pm$ 0.70 & 72.50 $\pm$ 0.70 & 79.00 $\pm$ 0.30 & 97.30 $\pm$ 0.20 \\
			\Rebuttal{DisenGCN \cite{ma2019disentangled}} & \Rebuttal{83.7} & \Rebuttal{73.2} & \Rebuttal{80.5} & \Rebuttal{-} \\
			VEPM \cite{he2022variational}& 84.3 $\pm$ 0.1 & 72.5 $\pm$ 0.1 & 82.4 $\pm$ 0.2& -\\
			\Rebuttal{IPGDN \cite{liu2020independence}} & \Rebuttal{84.1} & \Rebuttal{74.0} & \Rebuttal{81.2} & - \\
			\midrule
			MA-GCL \cite{gong2023ma} & 83.3 $\pm$ 0.4 & 73.6 $\pm$ 0.1 & 83.5 $\pm$ 0.4 & - \\
			MVGRL \cite{hassani2020contrastive} & 83.50 $\pm$ 0.40 & 73.30 $\pm$ 0.50 & 80.10 $\pm$ 0.70 & - \\
			InfoGCL \cite{xu2021infogcl} & 83.50 $\pm$ 0.30 & 73.50 $\pm$ 0.40 & 79.10 $\pm$ 0.20 & - \\
			DGI \cite{DBLP:conf/iclr/VelickovicFHLBH19} & 82.30 $\pm$ 0.60 & 71.80 $\pm$ 0.70 & 76.80 $\pm$ 0.60 & 63.80 $\pm$ 0.20 \\
			GRACE \cite{zhu2020deep} & 81.90 $\pm$ 0.40 & 71.20 $\pm$ 0.50 & 80.60 $\pm$ 0.40 & 69.71 $\pm$ 0.17 \\
			BGRL \cite{thakoor2021large} & 82.70 $\pm$ 0.60 & 71.10 $\pm$ 0.80 & 79.60 $\pm$ 0.50 & 73.63 $\pm$ 0.16 \\
			CCA-SSG \cite{zhang2021canonical} & 84.20 $\pm$ 0.40 & 73.10 $\pm$ 0.30 & 81.00 $\pm$ 0.40 & 73.34 $\pm$ 0.17 \\
			\Rebuttal{$\mathrm{SFA}_\mathrm{BT}$ \cite{zhang2023spectral}} & \Rebuttal{84.10 $\pm$ 0.01} & \Rebuttal{73.73 $\pm$ 0.03} & \Rebuttal{85.63 $\pm$ 0.07}& -  \\
			\midrule
			GAE \cite{kipf2016variational}& 71.50 $\pm$ 0.40 & 65.80 $\pm$ 0.40 & 72.10 $\pm$ 0.50 & - \\
			VGAE \cite{kipf2016variational} & 76.30 $\pm$ 0.20 & 66.80 $\pm$ 0.20 & 75.80 $\pm$ 0.40 & - \\
			Bandana \cite{zhao2024masked} & 84.62 $\pm$ 0.37 & 73.60 $\pm$ 0.16 & \textbf{83.53} $\pm$ 0.51 & - \\
			GiGaMAE \cite{shi2023gigamae}& 84.72 $\pm$ 0.47  & 72.31 $\pm$ 0.50  & -&-  \\
			SEEGERA \cite{li2023seegera} & 84.30 $\pm$ 0.40 & 73.00 $\pm$ 0.80 & 80.40 $\pm$ 0.40 & - \\
			GraphMAE \cite{hou2022graphmae} & 84.20 $\pm$ 0.40 & 73.40 $\pm$ 0.40 & 81.10 $\pm$ 0.40 & 74.50 $\pm$ 0.29 \\
			GraphMAE2 \cite{hou2023graphmae2} &  84.50 $\pm$ 0.60  &  73.40 $\pm$ 0.30  & 81.40 $\pm$ 0.50  & -\\
			\midrule
			\rowcolor{gray!30}
			\textbf{DiGGR}  & \textbf{84.96} $\pm$ 0.32 & \textbf{73.98} $\pm$ 0.27 & 81.30 $\pm$ 0.26 & \textbf{78.30} $\pm$ 0.71 \\
			\bottomrule
		\end{tabular}
	}
\end{table}

\paragraph{\Rebuttal{Node Classification for Heterophily Graph}}\label{subsubsec: node_clf_heter}
The baseline models for node classification on heterophilous graphs can be divided into two categories: \textit{i) Supervised methods}, which are specifically designed for non-homophily settings. These include GREET \cite{liu2023beyond}, MUSE \cite{yuan2023muse}, OGNN \cite{songordered2023}, LRGNN \cite{liang2023predicting} and LINKX \cite{yan2022two}. \textit{ii) Contrastive learning Methods}, which learn representations by maximizing mutual information across augmented views. This category includes GraphACL \cite{xiao2023simple} and DSSL \cite{zhu2021dssl}. We also ran the results of GraphMAE \cite{hou2022graphmae} only as a reference for comparision under the random masking strategy. 
\Rebuttal{The results are listed in Tab. \ref{tab: node_heter_clf}. DiGGR demonstrates competitive performance on these challenging benchmarks, achieving the best result on the Cornell and Actor, and comparable results on Texas and Wisconsin. This demonstrates that even when confronted with supervised and contrastive algorithms specifically designed for heterophilous scenarios, DiGGR can still achieve robust gains on limited-structure graphs through its trainable factorized graph decomposition and intra-factor GMAE reconstruction.}

\subsection{Graph Classification}
\label{subsec_4.2: graph class}
The backbone of encoder and decoder is GIN \cite{xu2018powerful}, commonly used in previous graph classification works. 
The evaluation protocol primarily follows GraphMAE \cite{hou2022graphmae}. Notice that we only utilize the factor-wise latent representation $H_d$ for the downstream task. Subsequently, we feed it into a downstream LIBSVM \cite{chang2011libsvm} classifier to predict the label and report the mean 10-fold cross-validation accuracy with standard deviation after 5 runs. We set the initial learning rate to 0.0005 with cosine learning rate decay for most cases. For the evaluation, the parameter C of SVM is searched in the set $\{10^{-3}, ..., 10\}$.
\Rebuttalmr{We categorized the baseline models into four groups:
$i)$ supervised methods, including GIN \cite{xu2018powerful}, DiffPool \cite{ying2018hierarchical}, etc.;
$ii)$ classical graph kernel methods, including WL \cite{shervashidze2011weisfeiler} and DGK \cite{yanardag2015deep};
$iii)$ contrastive learning methods, including GraphCL \cite{you2020graph}, MVGRL \cite{hassani2020contrastive}, etc.;
$iv)$ generative learning methods, including GraphMAE2\cite{hou2023graphmae2} and graph2vec \cite{narayanan2017graph2vec}, etc.}
Per graph classification research tradition, we report results from previous papers if available.

The graph classification results in Table \ref{table: Graph Classification} show that DiGGR achieves the best or competitive performance on most datasets, and its accuracy is comparable to supervised GNNs such as GIN and DiffPool (e.g., higher scores on IMDB-B and IMDB-M). Compared with random-masking generative baselines like GraphMAE, DiGGR yields consistent gains on IMDB-M, COLLAB and PROTEINS, indicating that factor-guided masking brings tangible benefits.

\subsection{Exploratory Studies}\label{subsec_4.4: explore}

\paragraph{Visualization and task-relevant factors}

\begin{figure}[t]
    \centering
    \begin{minipage}[b]{0.4\linewidth}
        \centering
        \includegraphics[width=\linewidth]{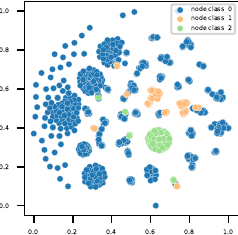}
        \caption*{(a) $z$, node label}
        \label{fig: mutag_z}
    \end{minipage}
    \hspace{1em}
    \begin{minipage}[b]{0.43\linewidth}
        \centering
        \includegraphics[width=\linewidth]{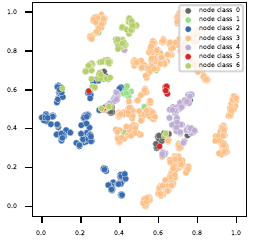}
        \caption*{(b) $H$, factor label}
        \label{fig: mutag_h}
    \end{minipage}
    \caption{T-SNE visualization of the MUTAG dataset, where $z$ is the latent factor, and $H$ is the learned node representation used for downstream tasks.}
    \label{fig: MUTAG_vis}
\end{figure}

\Rebuttalmr{To inspect how latent factors affect classification, we visualize the node–factor affiliation $z$ and the final node representation $H$ on MUTAG using 2D t-SNE (Fig. \ref{fig: MUTAG_vis}), colored by node label and factor label respectively. For clarity, only the three dominant node types are shown.} Nodes with the same label already exhibit clear clustering in the $z$-space, and the downstream representation $H$ becomes even more separable, indicating that DiGGR learns label-aware, factor-structured embeddings without supervision.
\Rebuttalmr{Complementary to the t-SNE plots, we follow \cite{he2022variational} to assess the statistical association between the learned latent factors and the downstream task by computing the Normalized Mutual Information (NMI) between factor assignments and ground-truth node labels.
} NMI is a metric that ranges from 0 to 1, where higher values signify more robust statistical dependencies between two random variables. In MUTAG datasets which comprises 7 distinct node types, the obtained NMI value was 0.5458. These results highlight that the latent factors obtained through self-supervised training are meaningful for the task, enhancing the correlation between the inferred latent factors and the task.

\begin{figure}[tbp]
    \centering
    \begin{minipage}[b]{0.48\linewidth}
        \centering
        \includegraphics[width=\linewidth]{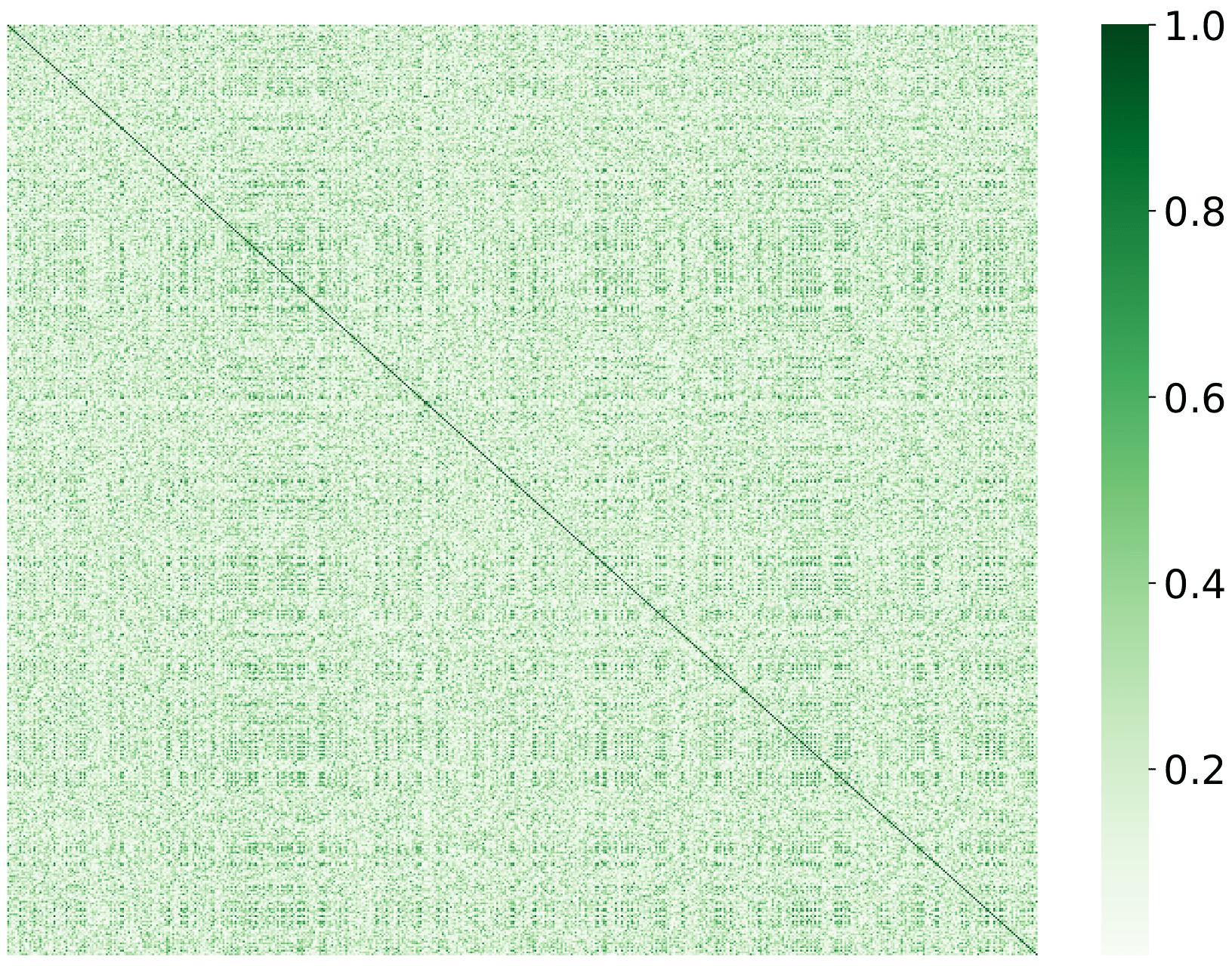}
        \caption*{(a) GraphMAE}
        \label{fig: Disg_cora_graphmae}
    \end{minipage}
    \hspace{1em}
    \begin{minipage}[b]{0.45\linewidth}
        \centering
        \includegraphics[width=\linewidth]{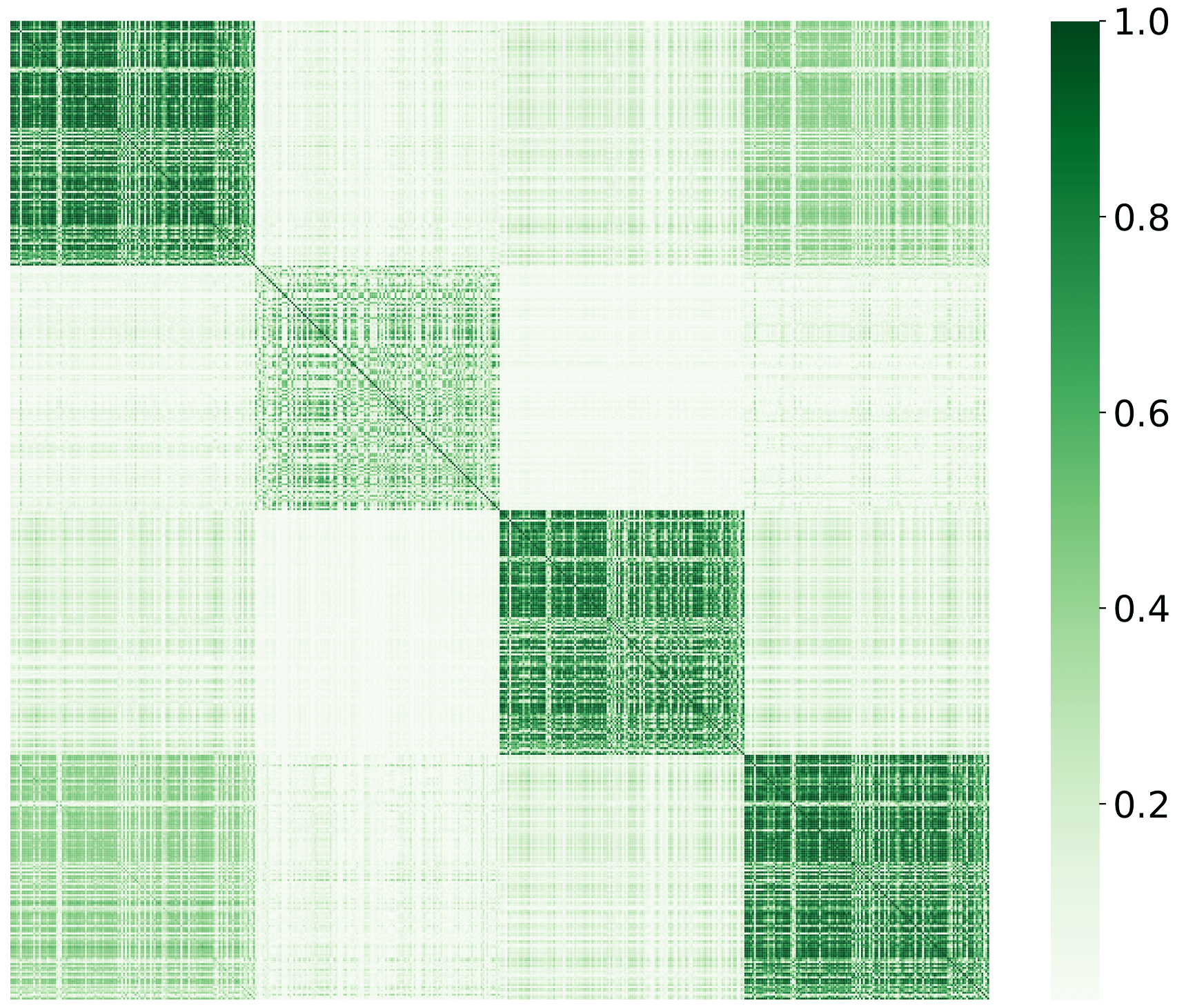}
        \caption*{(b) DiGGR}
        \label{fig: Disg_cora}
    \end{minipage}
    \caption{representation correlation matrix on Cora with number of factors $K = 4$. \textbf{\textit{(a)}} depicts the representation of entanglement, while \textbf{\textit{(b)}} illustrates disentanglement.}
    \label{fig: Disentangled_Cora}
\end{figure}

\paragraph{Disentangled representations}
To assess DiGGR's ability to disentangle representations for downstream tasks, we visualize the correlation structure of the learned node embeddings in Figure~\ref{fig: Disentangled_Cora}. The heatmaps show the absolute correlation between coordinates of the 512-dimensional graph representations obtained by GraphMAE and DiGGR. The representation from GraphMAE exhibits widespread correlations, whereas DiGGR produces a clearer block-like pattern, suggesting that different factors capture largely complementary information and that the hidden representation is partially disentangled.
\Rebuttal{\Rebuttalmr{We further adopt the quantitative disentanglement metric \textbf{SEPIN} \cite{do2019theory}, following \cite{wangnon2024}, with results reported in Table~\ref{tab_5: sepin@cora}.} DiGGR-MSE denotes the variant that replaces the proposed Scored Cosine Error (SCE) with mean squared error (MSE). DiGGR(SCE) consistently achieves higher SEPIN scores than DiGGR-MSE, and DiGGR outperforms GraphMAE across different top-$k$ dimensions with generally smaller variances, indicating both the effectiveness of SCE and the stronger feature-wise disentanglement of DiGGR.}

\begin{table}[tbp]
\centering
\caption{Feature disentanglement score (\textbf{SEPIN}) on Cora, where larger values indicate better disentanglement and @$k$ denotes the top-$k$ dimensions. Values are scaled by $10^3$ and we report mean$\pm$std over 5 seeds.}
\label{tab_5: sepin@cora}
{
\begin{tabular}{@{}cccc@{}}
\toprule
\textbf{Cora} & \textbf{SEPIN@1} & \textbf{SEPIN@10} & \textbf{SEPIN@all} \\ 
\midrule
GraphMAE & 3.79 $\pm$ 0.7014 & 2.58 $\pm$ 0.1938 & 0.67 $\pm$ 0.0178 \\
DiGGR-MSE   & 2.79 $\pm$ 0.0835 & 2.44 $\pm$ 0.0429 & 0.53 $\pm$ 0.0183 \\ 
\rowcolor{gray!30} DiGGR    & \textbf{3.98 $\pm$ 0.1674} & \textbf{2.94 $\pm$ 0.0448} & \textbf{0.71 $\pm$ 0.0039} \\ 
\bottomrule
\end{tabular}%
}%
\end{table}

\subsection{Ablation Studies}\label{subsec: ablation}
\subsubsection{Number of factor} 
\label{appendix_a.1.1: factor number}
\begin{figure}[tbp]
	\centering
	\begin{minipage}[b]{1.1\linewidth}
		\centering
		\begin{minipage}[b]{0.43\linewidth}
			\centering
			\includegraphics[width=\linewidth]{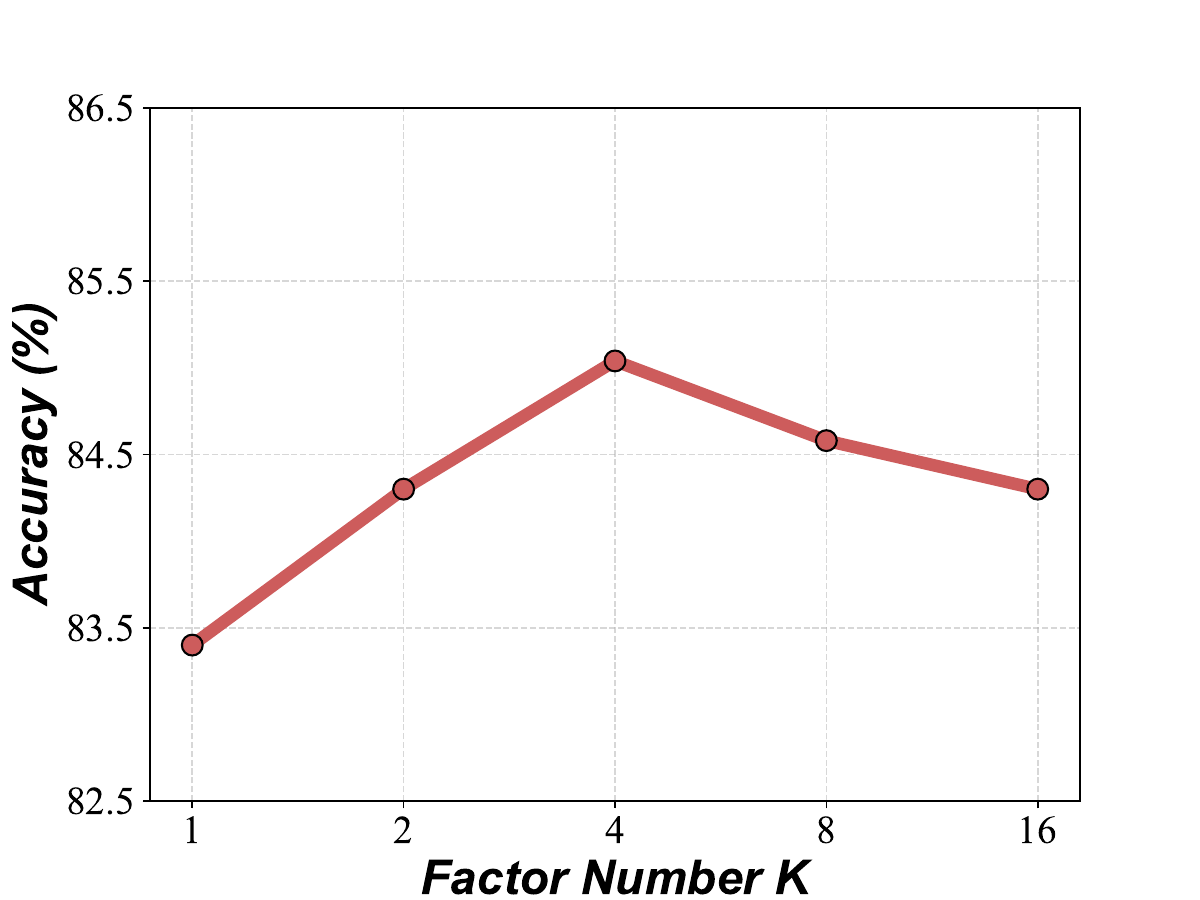}
			\caption*{(a) Cora}
			\label{fig:cora}
		\end{minipage}
		\begin{minipage}[b]{0.41\linewidth}
			\centering
			\includegraphics[width=\linewidth]{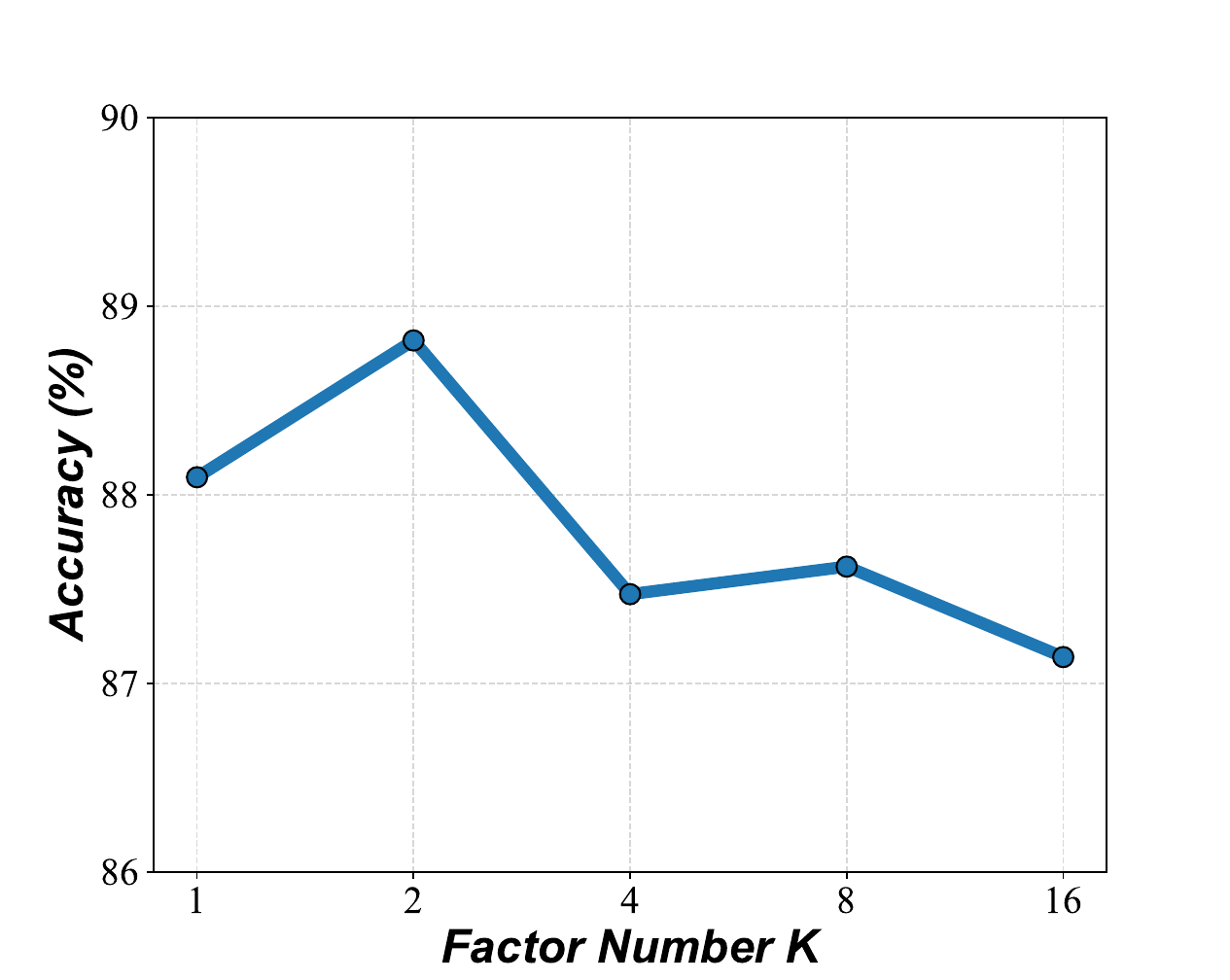}
			\caption*{(b) MUTAG}
			\label{fig:mutag}
		\end{minipage}
	\end{minipage}
	\caption{Performance of the task under different choices of latent factor number $K$, where the horizontal axis represents the change in $K$ and the vertical axis is accuracy.}
	\label{fig: ablation_K}
\end{figure}

\Rebuttalmr{A crucial hyperparameter in DiGGR is the number of latent factors $K$. When $K = 1$, DiGGR degenerates to GraphMAE with random masking over the whole graph.} As shown in Figure \ref{fig: ablation_K}, the best performance is typically achieved with $K$ in the range of 2–4, suggesting that $K$ should match dataset complexity. In practice, we use NMI to detect factor redundancy (high NMI indicates a redundant new factor). Meanwhile, DiGGR is relatively insensitive to $K$ over a wide range: varying $K$ from $1$ to $16$ yields only modest changes, where the standard deviation across $K$ is 0.58 on MUTAG and only 0.25 on Cora.

 \begin{table}[t]
 	\renewcommand{\arraystretch}{1.3}
 	\caption{The performance of the factor-wise GMAE and global-level GMAE on node classification (Cora, Citeseer) and graph classification (IMDB-B) tasks. For each model configuration, we conducted five independent runs, and the better results are highlighted in bold.}
 	\label{tab: branch_ablation}
 	{
 		\begin{tabular}{cccc}
 			\hline
 			& Cora             & Citeseer         & \multicolumn{1}{c}{IMDB-B} \\ \hline
 			graph-level GMAE & \textbf{84.04 $\pm$ 0.53} & \textbf{73.30 $\pm$ 0.38} & 74.79 $\pm$ 0.21                \\
 			factor-wise GMAE & 83.06 $\pm$ 0.26 & 71.60 $\pm$ 0.33 & \textbf{77.36 $\pm$ 0.51}                \\ \hline
 	\end{tabular}}
 \end{table}

\subsubsection{\Rebuttal{Representation and Branch Contributions}}
\paragraph{Representation for downstream tasks}
\label{appendix_a.1.2: ablation_repre}
\Rebuttalmr{We evaluate different combinations of $H_d$ and $H_g$ for downstream tasks (Table \ref{table_4.3:used_repre}). For node classification, concatenating $H_d$ and $H_g$ performs best, whereas for graph classification $H_d$ alone is sufficient. This is likely because graph pooling in graph classification already aggregates global context, while node classification uses unpooled node embeddings and thus benefits more from additional graph-level information $H_g$.}

\paragraph{\Rebuttal{Branch Ablation}}\label{subsubsec: ablation_branch}
\Rebuttalmr{We ablate the two branches by removing factor-wise GMAE or global-level GMAE during training. Specifically, we define variants in Eq. \ref{Eq_loss} by setting $\lambda_g = 0$ (factor-wise GMAE) or $\lambda_d = 0$ and $\lambda_z = 0$ (global-level GMAE), where removing factor-wise GMAE reduces DiGGR to a standard GMAE.} As shown in Table \ref{tab: branch_ablation}, removing either branch degrades performance across tasks. The global-level branch is more critical for node classification, while factor-wise GMAE remains relatively competitive for graph classification due to pooling, consistent with Table \ref{table_4.3:used_repre}. These results indicate that the two branches are complementary and their integration yields the best representations.




\subsubsection{\Rebuttal{Latent Fatcor Learning Module}}
\paragraph{Probabilistic Latent Factor Learning Module}
Although previous studies \cite{ma2019disentangled, li2021disentangled, li2022disentangled, wang2020disentangled, wang2022deep} have consistently demonstrated the efficacy of disentangled latent factor learning in graph contrastive learning, applying these methods directly to masked graph autoencoders introduces notable challenges. This issue is discussed in depth in Section \ref{F.a: disentangle_anylsis}.
To validate this observation, we replaced the latent factor learning module in DiGGR with a Non-Probabilistic Factor Learning strategy, wherein a node's hidden representation is mapped directly to a simplex point vector through a softmax function. 
We then performed ablations across datasets (Table \ref{table_4.3: disentangled}), where the non-probabilistic variant yields almost perfectly overlapping factors (NMI close to 1.0), while DiGGR reduces NMI to roughly one third of that value, confirming that the Gamma prior encourages more distinguishable latent factors and better convergence.

\paragraph{\Rebuttal{Effectiveness of Latent Factor Learning}}
\label{subsubsec: ablation_diggr_random}

\begin{table*}[t]
	\centering
	\caption{The NMI between the latent factors extracted by DiGGR and Non-probabilistic factor learning method across various datasets, and its performance improvement compared to GraphMAE, are examined. A lower NMI indicates a more pronounced disentanglement between factor-specific graphs, resulting in a greater performance enhancement.}
	\scalebox{1.1}{
		\begin{tabular}{@{}c|c|ccccccc@{}}
			\toprule
			\multicolumn{1}{l|}{}     & Dataset  & RDT-B & MUTAG & NCI-1 & IMDB-B & PROTEINS & COLLAB & IMDB-M \\ \midrule
			\multirow{2}{*}{DiGGR} & NMI      & 0.95  & 0.90  &  0.89  & 0.82   & 0.76     & 0.35   & 0.24       \\
			& ACC Gain &  + 0.18\%  &  + 0.53\% & + 0.83\%  &  + 2.16\%  & + 2.1\%  & + 3.44\% &  + 3.14\%       \\ \midrule
			\multirow{2}{*}{\begin{tabular}[c]{@{}c@{}}Non-probabilistic \\ Factor Learning\end{tabular}}     & NMI   & 1.00      & 1.00      &  0.80     &   1.00     & 0.60         &     1.00   &    0.94    \\
			& ACC Gain &    -2.23\%   & -2.02\%     &  -0.45\%     &   -0.80\%     &  -2.15\%       &  -3.00\%     & -0.11\%     \\ \bottomrule
	\end{tabular}}
	\label{table_4.3: disentangled}
\end{table*}

\begin{table}[tbp]
	\centering
	\renewcommand{\arraystretch}{1.1} 
	\caption{Ablation results comparing DiGGR against GraphMAE baselines and a random-partitioning variant (DiGGR-random), where factor groups are assigned randomly instead of being learned.}
	\label{tab: ablation_factor_random}
	\scalebox{1.1}{
		\begin{tabular}{cccc}
			\toprule
			Methods & IMDB-B &  Cora \\
			\midrule
			
			GraphMAE   & 75.52 $\pm$ 0.66 & 84.20 $\pm$ 0.40\\
			GraphMAE2   &73.88 $\pm$ 0.53 & 84.50 $\pm$ 0.60  \\
			\midrule
			DiGGR & \textbf{77.68} $\pm$  0.48 & \textbf{84.96} $\pm$  0.32 \\
			DiGGR-random & 75.65 $\pm$ 0.64 & 83.98 $\pm$ 0.28\\
			\bottomrule
		\end{tabular}
	}
\end{table}

\begin{table}[t]
	\centering
	\caption{Classification accuracy across datasets over 5 random seeds, using different combinations of representations ($H_d$ and $H_g$). The highest accuracy is highlighted in \textbf{bold}.}
	\scalebox{0.9}{\begin{tabular}{cc|cccc}
			\toprule
			$H_d$ & $H_g$ & Cora & IMDB-MULTI & Citeseer & PROTEINS\\ 
			\midrule 
			\checkmark & & 61.10 $\pm$ 1.83 & \textbf{54.77} $\pm$ 2.63 & 71.82 $\pm$ 0.98 & \textbf{77.76} $\pm$ 2.46 \\[2pt] 
			& \checkmark & 84.22 $\pm$ 0.38 & 51.62 $\pm$ 0.61 & 73.41 $\pm$ 0.43 & 75.52 $\pm$ 0.49 \\[2pt]  
			\checkmark & \checkmark & \textbf{84.96} $\pm$ 0.32 & 54.69 $\pm$ 2.06 & \textbf{73.98} $\pm$ 0.27 & 77.61 $\pm$ 0.12 \\ 
			\bottomrule
	\end{tabular}}
	\label{table_4.3:used_repre}
\end{table}

\Rebuttal{To isolate the contribution of disentanglement, we conducted a controlled experiment by replacing the probabilistic latent factor learning in DiGGR with a random partitioning strategy (denoted as DiGGR-random), which maintains the same subsequent GMAE architecture and fusion strategies.
The results (as shown in the Table \ref{tab: ablation_factor_random}) reveal a consistent and non-trivial performance gap between DiGGR-random and the full DiGGR across both datasets. This demonstrates that the gain of our framework stems not merely from the use of multiple representations, but fundamentally from the disentangling process itself. The random partitioning, while serving as a strong baseline that utilizes identical downstream components, fails to capture semantically meaningful factors, thus underscoring the necessity and advantage of our learnable, probabilistic disentanglement module.}

\begin{table}[t]
	\centering
	\renewcommand{\arraystretch}{1.2}
	\caption{Computational and inference efficiency on MUTAG: comparison of GFLOPs, end-to-end inference latency (P50/P90), and peak memory usage; lower values indicate better efficiency. The best results were highlighted in \textbf{bold}.}
	\scalebox{1.0}{
		\begin{tabular}{@{}lcccc@{}}
			\toprule
			\multicolumn{1}{c}{\textbf{MUTAG}} & \textbf{GFLOPs} & \textbf{P50 (ms)} & \textbf{P90 (ms)} & \textbf{Memory (MiB)} \\ \midrule
			DiGGR                              & 0.08            & 5.33              & 5.80              & 16.07                 \\
			GraphMAE                           & \textbf{0.03}            & 2.80            & 4.40              & \textbf{7.00}                  \\
			DisenSemi                          & 0.68            & 114.94            & 140.42            & 162.85                \\
			AutoGCL                            & 0.90            & \textbf{0.92}                  &  \textbf{1.35}                 & 61.39                 \\ \bottomrule
		\end{tabular}
		\label{tab_9: computation_analysis}
	}
\end{table}

\begin{table}[t]
	\centering
	\renewcommand{\arraystretch}{1.1}
	\caption{Results of DiGGR with approximate sampling on large-scale graph benchmarks remain competitive, confirming its robustness in scalable settings.}
	\label{tab: ogbn_1}
	{
		\scalebox{1.}{\begin{tabular}{lcc}
				\toprule
				& \textbf{ogbn-arxiv} & \textbf{ogbn-products} \\
				\midrule
				GraphMAE [2]  & 71.03 $\pm$ 0.02 & 78.89 $\pm$ 0.01 \\
				GraphMAE2 [3] & 71.89 $\pm$ 0.03 & \textbf{81.59 $\pm$ 0.02} \\
				\rowcolor{gray!30}	DiGGR         & \textbf{72.12 $\pm$ 0.08} & 79.13 $\pm$ 0.01 \\
				\bottomrule
	\end{tabular}}}
\end{table}

\subsubsection{\Rebuttal{Computation Analysis}}

\Rebuttal{Table \ref{tab_9: computation_analysis} compares the computational and deployment costs of DiGGR (our method, disentanglement-oriented and generative) with GraphMAE (generative), DisenSemi (disentanglement-oriented), and AutoGCL (contrastive) on the MUTAG graph-classification task. We report theoretical compute (GFLOPs), end-to-end inference latency (P50/P90) to characterize typical and tail delays, and peak GPU memory during inference. Overall, DiGGR exhibits substantially lower latency and memory footprint than DisenSemi, while remaining broadly comparable to GraphMAE and AutoGCL. These results suggest that DiGGR maintains competitive efficiency and is suitable for practical deployment.}
\Rebuttal{While DiGGR demonstrates a favorable balance between efficiency and performance on small-to-medium-scale graphs, its direct application to very large-scale graphs (e.g., ogbn-arxiv, ogbn-products) faces scalability challenges. This limitation primarily stems from the computational complexity associated with the graph edge likelihood calculation in Eq. \ref{eq: VI_3}. However, this is not a limitation unique to DiGGR, but rather a common challenge for probabilistic graph methods, as seen in \cite{li2023seegera, he2022variational, simonovsky2018graphvae}. To address this issue, we employ a stochastic subgraph decoding strategy \cite{salha2021fastgae}, which approximates edge likelihood computation by sampling subsets of the adjacency matrix, thereby enabling efficient training on large-scale graphs. The experimental configuration for these large-scale experiments was set as follows: Factor Number $K=4$, learning rate = 0.0005, and training Epochs = 600, while all other experimental settings remained consistent with those described in the main text.}

\Rebuttal{As shown in Table \ref{tab: ogbn_1}, despite employing approximate computation, DiGGR achieves highly competitive results on both benchmark datasets: it obtains optimal performance on ogbn-arxiv and comparable performance on ogbn-products. These results demonstrate the robustness of DiGGR to computational approximation and confirm that its disentanglement-guided masking mechanism remains effective under subgraph sampling. Consequently, the advantages of DiGGR are preserved in large-scale scenarios where computational constraints necessitate the adoption of approximation strategies.}

\subsection{Why DiGGR works better}
\label{F.a: disentangle_anylsis}
We conduct quantitative experiments to demonstrate that disentangled learning can effectively enhance the quality of representations learned by GMAE.
 The Normalized Mutual Information (NMI) is used to quantify the disentangling degree of different datasets. 
Generally, the NMI represents the similarity of node sets between different factor-specific graphs, and the \textit{lower NMI suggests a better-disentangled degree} with lower similarity among factor-specific graphs. 
The NMI between latent factors and the corresponding performance gain (compared to GraphMAE) is shown in the Table.\ref{table_4.3: disentangled}. It shows that DiGGR's performance improvement has a positive correlation with disentangled degree, where the better the disentangled degree, the more significant the performance improvement.
Methods employing Non-probabilistic Factor Learning often yield a NMI value approaching 1. This observation suggests that directly integrating Non-probabilistic Factor Learning into masked graph autoencoders hampers the discovery of meaningful factors, ultimately limiting performance improvements. The underlying issue stems from the difficulty faced by the factor learning module in achieving convergence, which prevents the effective differentiation of latent factors. As a result, entangled latent factors provide misleading guidance for representation learning, leading to suboptimal outcomes.
Moreover, these challenges may be exacerbated by differences in the underlying mechanisms of contrastive and generative self-supervised learning. Contrastive learning, which benefits from robust data augmentation techniques, tends to promote the convergence of latent factor models. In contrast, generative methods, which rely solely on masking strategies, may struggle to facilitate such convergence effectively, further contributing to the performance gap.

\section{Conclusion}
In this work, we propose \textbf{DiGGR} (\textbf{Di}sentangled \textbf{G}enerative \textbf{G}raph \textbf{R}epresentation Learning), a framework aimed at addressing the challenges of achieving disentangled representations within masked graph autoencoders. Unlike traditional approaches, which often fail to differentiate between overlapping factors in graph data, DiGGR leverages latent disentangled factors to guide the learning process, resulting in more robust embeddings that align closely with the intrinsic structure of graph data.
To achieve this, we introduce a two-step methodology. First, we employ a probabilistic graph generation model to factorize the graph into its disentangled latent factors, thereby capturing the underlying structures while separating distinct latent components. Second, we develop a Disentangled Graph Masked Autoencoder framework, which incorporates the disentangled factors into the representation learning process of masked graph autoencoders. This design allows the model to leverage the disentangled information effectively, leading to improved robustness and enhanced representational clarity.
Extensive experiments on 15 datasets have demonstrated the effectiveness of DiGGR on three downstream tasks.

\section*{Acknowledgments}
The authors would like to thank anonymous reviewers for their insightful feedback and also editors for their efforts.

\bibliography{TNNLS-2025-P-40307}

@inproceedings{huenhancing,
  title={Enhancing Uncertainty Estimation and Interpretability with Bayesian Non-negative Decision Layer},
  author={Hu, Xinyue and Duan, Zhibin and Chen, Bo and Zhou, Mingyuan},
year={2025},
  booktitle={The Thirteenth International Conference on Learning Representations}
}

@inproceedings{simonovsky2018graphvae,
  title={Graphvae: Towards generation of small graphs using variational autoencoders},
  author={Simonovsky, Martin and Komodakis, Nikos},
  booktitle={Artificial Neural Networks and Machine Learning--ICANN 2018: 27th International Conference on Artificial Neural Networks, Rhodes, Greece, October 4-7, 2018, Proceedings, Part I 27},
  pages={412--422},
  year={2018},
  organization={Springer}
}

@article{do2019theory,
	title={Theory and evaluation metrics for learning disentangled representations},
	author={Do, Kien and Tran, Truyen},
	journal={arXiv preprint arXiv:1908.09961},
	year={2019}
}

@inproceedings{songordered2023,
	title={Ordered GNN: Ordering Message Passing to Deal with Heterophily and Over-smoothing},
	author={Song, Yunchong and Zhou, Chenghu and Wang, Xinbing and Lin, Zhouhan},
	booktitle={The Eleventh International Conference on Learning Representations},
	year={2023}
}

@inproceedings{yan2022two,
	title={Two sides of the same coin: Heterophily and oversmoothing in graph convolutional neural networks},
	author={Yan, Yujun and Hashemi, Milad and Swersky, Kevin and Yang, Yaoqing and Koutra, Danai},
	booktitle={2022 IEEE International Conference on Data Mining (ICDM)},
	pages={1287--1292},
	year={2022},
	organization={IEEE}
}

@article{xiao2022decoupled,
  title={Decoupled self-supervised learning for graphs},
  author={Xiao, Teng and Chen, Zhengyu and Guo, Zhimeng and Zhuang, Zeyang and Wang, Suhang},
  journal={Advances in Neural Information Processing Systems},
  volume={35},
  pages={620--634},
  year={2022}
}

@article{kipf2016semi,
  title={Semi-supervised classification with graph convolutional networks},
  author={Kipf, Thomas N and Welling, Max},
  journal={arXiv preprint arXiv:1609.02907},
  year={2016}
}

@article{tan2022mgae,
  title={Mgae: Masked autoencoders for self-supervised learning on graphs},
  author={Tan, Qiaoyu and Liu, Ninghao and Huang, Xiao and Chen, Rui and Choi, Soo-Hyun and Hu, Xia},
  journal={arXiv preprint arXiv:2201.02534},
  year={2022}
}

@article{hinton1993autoencoders,
  title={Autoencoders, minimum description length and Helmholtz free energy},
  author={Hinton, Geoffrey E and Zemel, Richard},
  journal={Advances in neural information processing systems},
  volume={6},
  year={1993}
}

@article{higgins2017beta,
  title={beta-vae: Learning basic visual concepts with a constrained variational framework.},
  author={Higgins, Irina and Matthey, Loic and Pal, Arka and Burgess, Christopher P and Glorot, Xavier and Botvinick, Matthew M and Mohamed, Shakir and Lerchner, Alexander},
  journal={ICLR (Poster)},
  volume={3},
  year={2017}
}

@inproceedings{jiang2020psgan,
  title={Psgan: Pose and expression robust spatial-aware gan for customizable makeup transfer},
  author={Jiang, Wentao and Liu, Si and Gao, Chen and Cao, Jie and He, Ran and Feng, Jiashi and Yan, Shuicheng},
  booktitle={Proceedings of the IEEE/CVF Conference on Computer Vision and Pattern Recognition},
  pages={5194--5202},
  year={2020}
}

@article{mercatali2022symmetry,
  title={Symmetry-induced disentanglement on graphs},
  author={Mercatali, Giangiacomo and Freitas, Andr{\'e} and Garg, Vikas},
  journal={Advances in neural information processing systems},
  volume={35},
  pages={31497--31511},
  year={2022}
}

@article{li2022disentangled,
  title={Disentangled graph contrastive learning with independence promotion},
  author={Li, Haoyang and Zhang, Ziwei and Wang, Xin and Zhu, Wenwu},
  journal={IEEE Transactions on Knowledge and Data Engineering},
  volume={35},
  number={8},
  pages={7856--7869},
  year={2022},
  publisher={IEEE}
}

@inproceedings{hou2023graphmae2,
  title={Graphmae2: A decoding-enhanced masked self-supervised graph learner},
  author={Hou, Zhenyu and He, Yufei and Cen, Yukuo and Liu, Xiao and Dong, Yuxiao and Kharlamov, Evgeny and Tang, Jie},
  booktitle={Proceedings of the ACM Web Conference 2023},
  pages={737--746},
  year={2023}
}

@inproceedings{shi2023gigamae,
  title={Gigamae: Generalizable graph masked autoencoder via collaborative latent space reconstruction},
  author={Shi, Yucheng and Dong, Yushun and Tan, Qiaoyu and Li, Jundong and Liu, Ninghao},
  booktitle={Proceedings of the 32nd ACM International Conference on Information and Knowledge Management},
  pages={2259--2269},
  year={2023}
}

@article{liang2023predicting,
	title={Predicting global label relationship matrix for graph neural networks under heterophily},
	author={Liang, Langzhang and Hu, Xiangjing and Xu, Zenglin and Song, Zixing and King, Irwin},
	journal={Advances in Neural Information Processing Systems},
	volume={36},
	pages={10909--10921},
	year={2023}
}

@inproceedings{zhu2021dssl,
	title={Dssl: Deep surroundings-person separation learning for text-based person retrieval},
	author={Zhu, Aichun and Wang, Zijie and Li, Yifeng and Wan, Xili and Jin, Jing and Wang, Tian and Hu, Fangqiang and Hua, Gang},
	booktitle={Proceedings of the 29th ACM international conference on multimedia},
	pages={209--217},
	year={2021}
}

@article{zhao2024masked,
  title={Masked Graph Autoencoder with Non-discrete Bandwidths},
  author={Zhao, Ziwen and Li, Yuhua and Zou, Yixiong and Tang, Jiliang and Li, Ruixuan},
  journal={arXiv preprint arXiv:2402.03814},
  year={2024}
}

@inproceedings{li2023seegera,
  title={Seegera: Self-supervised semi-implicit graph variational auto-encoders with masking},
  author={Li, Xiang and Ye, Tiandi and Shan, Caihua and Li, Dongsheng and Gao, Ming},
  booktitle={Proceedings of the ACM web conference 2023},
  pages={143--153},
  year={2023}
}

@inproceedings{li2023s,
  title={What's Behind the Mask: Understanding Masked Graph Modeling for Graph Autoencoders},
  author={Li, Jintang and Wu, Ruofan and Sun, Wangbin and Chen, Liang and Tian, Sheng and Zhu, Liang and Meng, Changhua and Zheng, Zibin and Wang, Weiqiang},
  booktitle={Proceedings of the 29th ACM SIGKDD Conference on Knowledge Discovery and Data Mining},
  pages={1268--1279},
  year={2023}
}

@article{velickovic2017graph,
  title={Graph attention networks},
  author={Velickovic, Petar and Cucurull, Guillem and Casanova, Arantxa and Romero, Adriana and Lio, Pietro and Bengio, Yoshua and others},
  journal={stat},
  volume={1050},
  number={20},
  pages={10--48550},
  year={2017}
}

@article{hamilton2017inductive,
  title={Inductive representation learning on large graphs},
  author={Hamilton, Will and Ying, Zhitao and Leskovec, Jure},
  journal={Advances in neural information processing systems},
  volume={30},
  year={2017}
}

@inproceedings{yang2016revisiting,
  title={Revisiting semi-supervised learning with graph embeddings},
  author={Yang, Zhilin and Cohen, William and Salakhudinov, Ruslan},
  booktitle={International conference on machine learning},
  pages={40--48},
  year={2016},
  organization={PMLR}
}

@inproceedings{kakogeorgiou2022hide,   
title={What to hide from your students: Attention-guided masked image modeling}, author={Kakogeorgiou, Ioannis and Gidaris, Spyros and Psomas, Bill and Avrithis, Yannis and Bursuc, Andrei and Karantzalos, Konstantinos and Komodakis, Nikos},   booktitle={European Conference on Computer Vision},   
pages={300--318},   
year={2022},   
organization={Springer} }

@inproceedings{hou2022graphmae,
  title={Graph{MAE}: Self-supervised masked graph autoencoders},
  author={Hou, Zhenyu and Liu, Xiao and Cen, Yukuo and Dong, Yuxiao and Yang, Hongxia and Wang, Chunjie and Tang, Jie},
  booktitle={Proceedings of the 28th ACM SIGKDD Conference on Knowledge Discovery and Data Mining},
  pages={594--604},
  year={2022}
}

@inproceedings{mo2023disentangled,
  title={Disentangled multiplex graph representation learning},
  author={Mo, Yujie and Lei, Yajie and Shen, Jialie and Shi, Xiaoshuang and Shen, Heng Tao and Zhu, Xiaofeng},
  booktitle={International Conference on Machine Learning},
  pages={24983--25005},
  year={2023},
  organization={PMLR}
}

@article{bengio2013representation,
  title={Representation learning: A review and new perspectives},
  author={Bengio, Yoshua and Courville, Aaron and Vincent, Pascal},
  journal={IEEE transactions on pattern analysis and machine intelligence},
  volume={35},
  number={8},
  pages={1798--1828},
  year={2013},
  publisher={IEEE}
}

@inproceedings{he2022masked,
  title={Masked autoencoders are scalable vision learners},
  author={He, Kaiming and Chen, Xinlei and Xie, Saining and Li, Yanghao and Doll{\'a}r, Piotr and Girshick, Ross},
  booktitle={Proceedings of the IEEE/CVF conference on computer vision and pattern recognition},
  pages={16000--16009},
  year={2022}
}

@inproceedings{ma2019disentangled,
  title={Disentangled graph convolutional networks},
  author={Ma, Jianxin and Cui, Peng and Kuang, Kun and Wang, Xin and Zhu, Wenwu},
  booktitle={International conference on machine learning},
  pages={4212--4221},
  year={2019},
  organization={PMLR}
}

@article{gao2024hypergraph,
	title={Hypergraph computation},
	author={Gao, Yue and Ji, Shuyi and Han, Xiangmin and Dai, Qionghai},
	journal={Engineering},
	volume={40},
	pages={188--201},
	year={2024},
	publisher={Elsevier}
}

@article{cao2024immense,
	title={The immense impact of reverse edges on large hierarchical networks},
	author={Cao, Haosen and Hu, Bin-Bin and Mo, Xiaoyu and Chen, Duxin and Gao, Jianxi and Yuan, Ye and Chen, Guanrong and Vicsek, Tam{\'a}s and Guan, Xiaohong and Zhang, Hai-Tao},
	journal={Engineering},
	volume={36},
	pages={240--249},
	year={2024},
	publisher={Elsevier}
}

@article{li2021disentangled,
  title={Disentangled contrastive learning on graphs},
  author={Li, Haoyang and Wang, Xin and Zhang, Ziwei and Yuan, Zehuan and Li, Hang and Zhu, Wenwu},
  journal={Advances in Neural Information Processing Systems},
  volume={34},
  pages={21872--21884},
  year={2021}
}

@article{devlin2018bert,
  title={Bert: Pre-training of deep bidirectional transformers for language understanding},
  author={Devlin, Jacob and Chang, Ming-Wei and Lee, Kenton and Toutanova, Kristina},
  journal={arXiv preprint arXiv:1810.04805},
  year={2018}
}

@article{you2020graph,
  title={Graph contrastive learning with augmentations},
  author={You, Yuning and Chen, Tianlong and Sui, Yongduo and Chen, Ting and Wang, Zhangyang and Shen, Yang},
  journal={Advances in neural information processing systems},
  volume={33},
  pages={5812--5823},
  year={2020}
}

@inproceedings{wang2020disentangled,
  title={Disentangled graph collaborative filtering},
  author={Wang, Xiang and Jin, Hongye and Zhang, An and He, Xiangnan and Xu, Tong and Chua, Tat-Seng},
  booktitle={Proceedings of the 43rd international ACM SIGIR conference on research and development in information retrieval},
  pages={1001--1010},
  year={2020}
}

@article{wang2022deep,
  title={Deep generative model for periodic graphs},
  author={Wang, Shiyu and Guo, Xiaojie and Zhao, Liang},
  journal={Advances in Neural Information Processing Systems},
  volume={35},
  pages={35797--35810},
  year={2022}
}

@inproceedings{chen2020simple,
  title={A simple framework for contrastive learning of visual representations},
  author={Chen, Ting and Kornblith, Simon and Norouzi, Mohammad and Hinton, Geoffrey},
  booktitle={International conference on machine learning},
  pages={1597--1607},
  year={2020},
  organization={PMLR}
}

@inproceedings{zhou2015infinite,
  title={Infinite edge partition models for overlapping community detection and link prediction},
  author={Zhou, Mingyuan},
  booktitle={Artificial intelligence and statistics},
  pages={1135--1143},
  year={2015},
  organization={PMLR}
}

@article{zhang2018whai,
  title={WHAI: Weibull hybrid autoencoding inference for deep topic modeling},
  author={Zhang, Hao and Chen, Bo and Guo, Dandan and Zhou, Mingyuan},
  journal={arXiv preprint arXiv:1803.01328},
  year={2018}
}

@article{he2022variational,
  title={A Variational Edge Partition Model for Supervised Graph Representation Learning},
  author={He, Yilin and Wang, Chaojie and Zhang, Hao and Chen, Bo and Zhou, Mingyuan},
  journal={Advances in Neural Information Processing Systems},
  volume={35},
  pages={12339--12351},
  year={2022}
}

@article{ying2018hierarchical,
  title={Hierarchical graph representation learning with differentiable pooling},
  author={Ying, Zhitao and You, Jiaxuan and Morris, Christopher and Ren, Xiang and Hamilton, Will and Leskovec, Jure},
  journal={Advances in neural information processing systems},
  volume={31},
  year={2018}
}

@inproceedings{lin2017focal,
  title={Focal loss for dense object detection},
  author={Lin, Tsung-Yi and Goyal, Priya and Girshick, Ross and He, Kaiming and Doll{\'a}r, Piotr},
  booktitle={Proceedings of the IEEE international conference on computer vision},
  pages={2980--2988},
  year={2017}
}

@article{xu2018powerful,
  title={How powerful are graph neural networks?},
  author={Xu, Keyulu and Hu, Weihua and Leskovec, Jure and Jegelka, Stefanie},
  journal={arXiv preprint arXiv:1810.00826},
  year={2018}
}

@article{chang2011libsvm,
  title={LIBSVM: a library for support vector machines},
  author={Chang, Chih-Chung and Lin, Chih-Jen},
  journal={ACM transactions on intelligent systems and technology (TIST)},
  volume={2},
  number={3},
  pages={1--27},
  year={2011},
  publisher={Acm New York, NY, USA}
}

@article{hu2019strategies,
  title={Strategies for pre-training graph neural networks},
  author={Hu, Weihua and Liu, Bowen and Gomes, Joseph and Zitnik, Marinka and Liang, Percy and Pande, Vijay and Leskovec, Jure},
  journal={arXiv preprint arXiv:1905.12265},
  year={2019}
}

@article{tu2023rare,
  title={RARE: Robust Masked Graph Autoencoder},
  author={Tu, Wenxuan and Liao, Qing and Zhou, Sihang and Peng, Xin and Ma, Chuan and Liu, Zhe and Liu, Xinwang and Cai, Zhiping},
  journal={arXiv preprint arXiv:2304.01507},
  year={2023}
}

@inproceedings{tian2023heterogeneous,
  title={Heterogeneous graph masked autoencoders},
  author={Tian, Yijun and Dong, Kaiwen and Zhang, Chunhui and Zhang, Chuxu and Chawla, Nitesh V},
  booktitle={Proceedings of the AAAI Conference on Artificial Intelligence},
  volume={37},
  number={8},
  pages={9997--10005},
  year={2023}
}

@article{bao2021beit,
  title={Beit: Bert pre-training of image transformers},
  author={Bao, Hangbo and Dong, Li and Piao, Songhao and Wei, Furu},
  journal={arXiv preprint arXiv:2106.08254},
  year={2021}
}

@inproceedings{hassani2020contrastive,
  title={Contrastive multi-view representation learning on graphs},
  author={Hassani, Kaveh and Khasahmadi, Amir Hosein},
  booktitle={International conference on machine learning},
  pages={4116--4126},
  year={2020},
  organization={PMLR}
}

@article{zhu2020deep,
  title={Deep graph contrastive representation learning},
  author={Zhu, Yanqiao and Xu, Yichen and Yu, Feng and Liu, Qiang and Wu, Shu and Wang, Liang},
  journal={arXiv preprint arXiv:2006.04131},
  year={2020}
}

@article{thakoor2021large,
  title={Large-scale representation learning on graphs via bootstrapping},
  author={Thakoor, Shantanu and Tallec, Corentin and Azar, Mohammad Gheshlaghi and Azabou, Mehdi and Dyer, Eva L and Munos, Remi and Veli{\v{c}}kovi{\'c}, Petar and Valko, Michal},
  journal={arXiv preprint arXiv:2102.06514},
  year={2021}
}

@article{zhang2022mask,
  title={How mask matters: Towards theoretical understandings of masked autoencoders},
  author={Zhang, Qi and Wang, Yifei and Wang, Yisen},
  journal={Advances in Neural Information Processing Systems},
  volume={35},
  pages={27127--27139},
  year={2022}
}

@article{xu2021infogcl,
  title={Infogcl: Information-aware graph contrastive learning},
  author={Xu, Dongkuan and Cheng, Wei and Luo, Dongsheng and Chen, Haifeng and Zhang, Xiang},
  journal={Advances in Neural Information Processing Systems},
  volume={34},
  pages={30414--30425},
  year={2021}
}

@article{tsai2020self,
  title={Self-supervised learning from a multi-view perspective},
  author={Tsai, Yao-Hung Hubert and Wu, Yue and Salakhutdinov, Ruslan and Morency, Louis-Philippe},
  journal={arXiv preprint arXiv:2006.05576},
  year={2020}
}

@article{tian2020makes,
  title={What makes for good views for contrastive learning?},
  author={Tian, Yonglong and Sun, Chen and Poole, Ben and Krishnan, Dilip and Schmid, Cordelia and Isola, Phillip},
  journal={Advances in neural information processing systems},
  volume={33},
  pages={6827--6839},
  year={2020}
}

@article{zhang2021canonical,
  title={From canonical correlation analysis to self-supervised graph neural networks},
  author={Zhang, Hengrui and Wu, Qitian and Yan, Junchi and Wipf, David and Yu, Philip S},
  journal={Advances in Neural Information Processing Systems},
  volume={34},
  pages={76--89},
  year={2021}
}

@article{kipf2016variational,
  title={Variational graph auto-encoders},
  author={Kipf, Thomas N and Welling, Max},
  journal={arXiv preprint arXiv:1611.07308},
  year={2016}
}

@article{shervashidze2011weisfeiler,
  title={Weisfeiler-lehman graph kernels.},
  author={Shervashidze, Nino and Schweitzer, Pascal and Van Leeuwen, Erik Jan and Mehlhorn, Kurt and Borgwardt, Karsten M},
  journal={Journal of Machine Learning Research},
  volume={12},
  number={9},
  year={2011}
}

@inproceedings{yanardag2015deep,
  title={Deep graph kernels},
  author={Yanardag, Pinar and Vishwanathan, SVN},
  booktitle={Proceedings of the 21th ACM SIGKDD international conference on knowledge discovery and data mining},
  pages={1365--1374},
  year={2015}
}

@inproceedings{qiu2020gcc,
  title={Gcc: Graph contrastive coding for graph neural network pre-training},
  author={Qiu, Jiezhong and Chen, Qibin and Dong, Yuxiao and Zhang, Jing and Yang, Hongxia and Ding, Ming and Wang, Kuansan and Tang, Jie},
  booktitle={Proceedings of the 26th ACM SIGKDD international conference on knowledge discovery \& data mining},
  pages={1150--1160},
  year={2020}
}

@article{narayanan2017graph2vec,
  title={graph2vec: Learning distributed representations of graphs},
  author={Narayanan, Annamalai and Chandramohan, Mahinthan and Venkatesan, Rajasekar and Chen, Lihui and Liu, Yang and Jaiswal, Shantanu},
  journal={arXiv preprint arXiv:1707.05005},
  year={2017}
}

@article{sun2019infograph,
  title={Infograph: Unsupervised and semi-supervised graph-level representation learning via mutual information maximization},
  author={Sun, Fan-Yun and Hoffmann, Jordan and Verma, Vikas and Tang, Jian},
  journal={arXiv preprint arXiv:1908.01000},
  year={2019}
}

@inproceedings{you2021graph,
  title={Graph contrastive learning automated},
  author={You, Yuning and Chen, Tianlong and Shen, Yang and Wang, Zhangyang},
  booktitle={International Conference on Machine Learning},
  pages={12121--12132},
  year={2021},
  organization={PMLR}
}

@inproceedings{duan2021sawtooth,
  title={Sawtooth factorial topic embeddings guided gamma belief network},
  author={Duan, Zhibin and Wang, Dongsheng and Chen, Bo and Wang, Chaojie and Chen, Wenchao and Li, Yewen and Ren, Jie and Zhou, Mingyuan},
  booktitle={International Conference on Machine Learning},
  pages={2903--2913},
  year={2021},
  organization={PMLR}
}

@article{DBLP:journals/tnn/LiCZLHW24,
  author       = {Haifeng Li and
                  Jun Cao and
                  Jiawei Zhu and
                  Qinyao Luo and
                  Silu He and
                  Xuying Wang},
  title        = {Augmentation-Free Graph Contrastive Learning of Invariant-Discriminative
                  Representations},
  journal      = {{IEEE} Trans. Neural Networks Learn. Syst.},
  volume       = {35},
  number       = {8},
  pages        = {11157--11167},
  year         = {2024},
  url          = {https://doi.org/10.1109/TNNLS.2023.3248871},
  doi          = {10.1109/TNNLS.2023.3248871},
  bibsource    = {dblp computer science bibliography, https://dblp.org}
}

@inproceedings{DBLP:conf/www/XiaWCHL22,
  author       = {Jun Xia and
                  Lirong Wu and
                  Jintao Chen and
                  Bozhen Hu and
                  Stan Z. Li},
  editor       = {Fr{\'{e}}d{\'{e}}rique Laforest and
                  Rapha{\"{e}}l Troncy and
                  Elena Simperl and
                  Deepak Agarwal and
                  Aristides Gionis and
                  Ivan Herman and
                  Lionel M{\'{e}}dini},
  title        = {SimGRACE: {A} Simple Framework for Graph Contrastive Learning without
                  Data Augmentation},
  booktitle    = {{WWW} '22: The {ACM} Web Conference 2022, Virtual Event, Lyon, France,
                  April 25 - 29, 2022},
  pages        = {1070--1079},
  publisher    = {{ACM}},
  year         = {2022},
  url          = {https://doi.org/10.1145/3485447.3512156},
  doi          = {10.1145/3485447.3512156},
  bibsource    = {dblp computer science bibliography, https://dblp.org}
}

@inproceedings{DBLP:conf/iclr/HjelmFLGBTB19,
  author       = {R. Devon Hjelm and
                  Alex Fedorov and
                  Samuel Lavoie{-}Marchildon and
                  Karan Grewal and
                  Philip Bachman and
                  Adam Trischler and
                  Yoshua Bengio},
  title        = {Learning deep representations by mutual information estimation and
                  maximization},
  booktitle    = {7th International Conference on Learning Representations, {ICLR} 2019,
                  New Orleans, LA, USA, May 6-9, 2019},
  publisher    = {OpenReview.net},
  year         = {2019},
  url          = {https://openreview.net/forum?id=Bklr3j0cKX},
  bibsource    = {dblp computer science bibliography, https://dblp.org}
}

@inproceedings{gong2023ma,
  title={Ma-gcl: Model augmentation tricks for graph contrastive learning},
  author={Gong, Xumeng and Yang, Cheng and Shi, Chuan},
  booktitle={Proceedings of the AAAI Conference on Artificial Intelligence},
  volume={37},
  number={4},
  pages={4284--4292},
  year={2023}
}

@inproceedings{DBLP:conf/iclr/VelickovicFHLBH19,
  author       = {Petar Velickovic and
                  William Fedus and
                  William L. Hamilton and
                  Pietro Li{\`{o}} and
                  Yoshua Bengio and
                  R. Devon Hjelm},
  title        = {Deep Graph Infomax},
  booktitle    = {7th International Conference on Learning Representations, {ICLR} 2019,
                  New Orleans, LA, USA, May 6-9, 2019},
  publisher    = {OpenReview.net},
  year         = {2019},
  url          = {https://openreview.net/forum?id=rklz9iAcKQ},
  bibsource    = {dblp computer science bibliography, https://dblp.org}
}

@article{wang2012nonnegative,
  title={Nonnegative matrix factorization: A comprehensive review},
  author={Wang, Yu-Xiong and Zhang, Yu-Jin},
  journal={IEEE Transactions on knowledge and data engineering},
  volume={25},
  number={6},
  pages={1336--1353},
  year={2012},
  publisher={IEEE}
}

@inproceedings{khemakhem2020variational,
  title={Variational autoencoders and nonlinear ica: A unifying framework},
  author={Khemakhem, Ilyes and Kingma, Diederik and Monti, Ricardo and Hyvarinen, Aapo},
  booktitle={International conference on artificial intelligence and statistics},
  pages={2207--2217},
  year={2020},
  organization={PMLR}
}

@inproceedings{yin2022autogcl,
  title={Autogcl: Automated graph contrastive learning via learnable view generators},
  author={Yin, Yihang and Wang, Qingzhong and Huang, Siyu and Xiong, Haoyi and Zhang, Xiang},
  booktitle={Proceedings of the AAAI conference on artificial intelligence},
  volume={36},
  number={8},
  pages={8892--8900},
  year={2022}
}

@inproceedings{liu2023beyond,
	title={Beyond smoothing: Unsupervised graph representation learning with edge heterophily discriminating},
	author={Liu, Yixin and Zheng, Yizhen and Zhang, Daokun and Lee, Vincent CS and Pan, Shirui},
	booktitle={Proceedings of the AAAI conference on artificial intelligence},
	volume={37},
	number={4},
	pages={4516--4524},
	year={2023}
}

@article{xiao2023simple,
	title={Simple and asymmetric graph contrastive learning without augmentations},
	author={Xiao, Teng and Zhu, Huaisheng and Chen, Zhengyu and Wang, Suhang},
	journal={Advances in neural information processing systems},
	volume={36},
	pages={16129--16152},
	year={2023}
}

@article{salha2021fastgae,
	title={FastGAE: Scalable graph autoencoders with stochastic subgraph decoding},
	author={Salha, Guillaume and Hennequin, Romain and Remy, Jean-Baptiste and Moussallam, Manuel and Vazirgiannis, Michalis},
	journal={Neural Networks},
	volume={142},
	pages={1--19},
	year={2021},
	publisher={Elsevier}
}

@inproceedings{yuan2023muse,
	title={Muse: multi-view contrastive learning for heterophilic graphs},
	author={Yuan, Mengyi and Chen, Minjie and Li, Xiang},
	booktitle={Proceedings of the 32nd ACM International Conference on Information and Knowledge Management},
	pages={3094--3103},
	year={2023}
}

@inproceedings{zhang2023spectral,
  title={Spectral feature augmentation for graph contrastive learning and beyond},
  author={Zhang, Yifei and Zhu, Hao and Song, Zixing and Koniusz, Piotr and King, Irwin},
  booktitle={Proceedings of the AAAI conference on artificial intelligence},
  volume={37},
  number={9},
  pages={11289--11297},
  year={2023}
}

@inproceedings{liu2020independence,
  title={Independence promoted graph disentangled networks},
  author={Liu, Yanbei and Wang, Xiao and Wu, Shu and Xiao, Zhitao},
  booktitle={Proceedings of the AAAI Conference on Artificial Intelligence},
  volume={34},
  number={04},
  pages={4916--4923},
  year={2020}
}

@inproceedings{wangnon2024,
  title={Non-negative Contrastive Learning},
  author={Wang, Yifei and Zhang, Qi and Guo, Yaoyu and Wang, Yisen},
  year={2024},
  booktitle={The Twelfth International Conference on Learning Representations}
}

@inproceedings{li2022finding,
	title={Finding global homophily in graph neural networks when meeting heterophily},
	author={Li, Xiang and Zhu, Renyu and Cheng, Yao and Shan, Caihua and Luo, Siqiang and Li, Dongsheng and Qian, Weining},
	booktitle={International conference on machine learning},
	pages={13242--13256},
	year={2022},
	organization={PMLR}
}

@inproceedings{arora2013practical,
  title={A practical algorithm for topic modeling with provable guarantees},
  author={Arora, Sanjeev and Ge, Rong and Halpern, Yonatan and Mimno, David and Moitra, Ankur and Sontag, David and Wu, Yichen and Zhu, Michael},
  booktitle={International conference on machine learning},
  pages={280--288},
  year={2013},
  organization={PMLR}
}

@inproceedings{pei2020geom,
	title={GEOM-GCN: GEOMETRIC GRAPH CONVOLUTIONAL NETWORKS},
	author={Pei, Hongbin and Wei, Bingzhe and Chang, Kevin Chen Chuan and Lei, Yu and Yang, Bo},
	booktitle={8th International Conference on Learning Representations, ICLR 2020},
	year={2020}
}

@article{wang2024disensemi,
	title={DisenSemi: Semi-supervised graph classification via disentangled representation learning},
	author={Wang, Yifan and Luo, Xiao and Chen, Chong and Hua, Xian-Sheng and Zhang, Ming and Ju, Wei},
	journal={IEEE Transactions on Neural Networks and Learning Systems},
	year={2024},
	publisher={IEEE}
}

@inproceedings{locatello2019challenging,
  title={Challenging common assumptions in the unsupervised learning of disentangled representations},
  author={Locatello, Francesco and Bauer, Stefan and Lucic, Mario and Raetsch, Gunnar and Gelly, Sylvain and Sch{\"o}lkopf, Bernhard and Bachem, Olivier},
  booktitle={international conference on machine learning},
  pages={4114--4124},
  year={2019},
  organization={PMLR}
}

@article{gillis2023partial,
  title={Partial identifiability for nonnegative matrix factorization},
  author={Gillis, Nicolas and Rajk{\'o}, R{\'o}bert},
  journal={SIAM Journal on Matrix Analysis and Applications},
  volume={44},
  number={1},
  pages={27--52},
  year={2023},
  publisher={SIAM}
}
\bibliographystyle{IEEEtran}

	\begin{IEEEbiography}[{\includegraphics[width=1in,height=1.25in,clip,keepaspectratio]{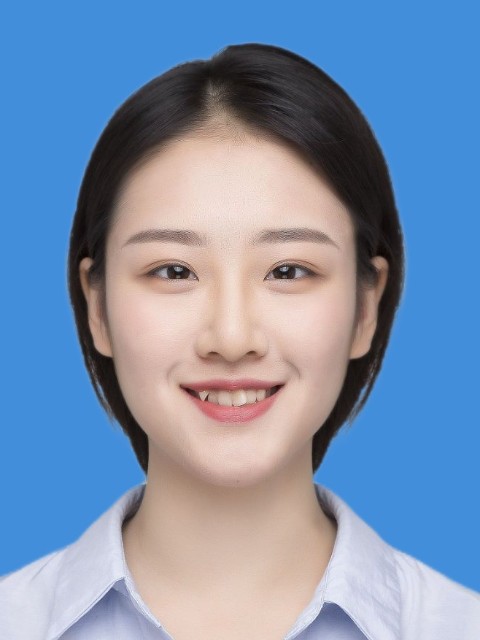}}]{Xinyue Hu}
	Xinyue Hu received the  B.S. degree  in engineering from Capital Normal University, Beijing, China, in 2022. She is currently pursuing the Ph.D. degree at Xidian University, Xi’an, China. Her research interests include deep learning and statistical machine learning.
\end{IEEEbiography}	
\vspace{-5mm}
\begin{IEEEbiography}[{\includegraphics[width=1in,height=1.25in,clip,keepaspectratio]{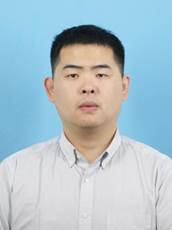}}]{Zhibin Duan}
	received  BS degree from Xidian University, Xi’an, China, in 2019, and PhD degree in electronic engineering from the same university in 2024.
	He is currently an Assistant Professor at School of Mathematics and Statistics, Xi’an Jiaotong University, Xi’an, Shaanxi, China. 
	His research interests include Bayesian deep learning.
\end{IEEEbiography}
\vspace{-5mm}
\begin{IEEEbiography}[{\includegraphics[width=1in,height=1.25in,clip,keepaspectratio]{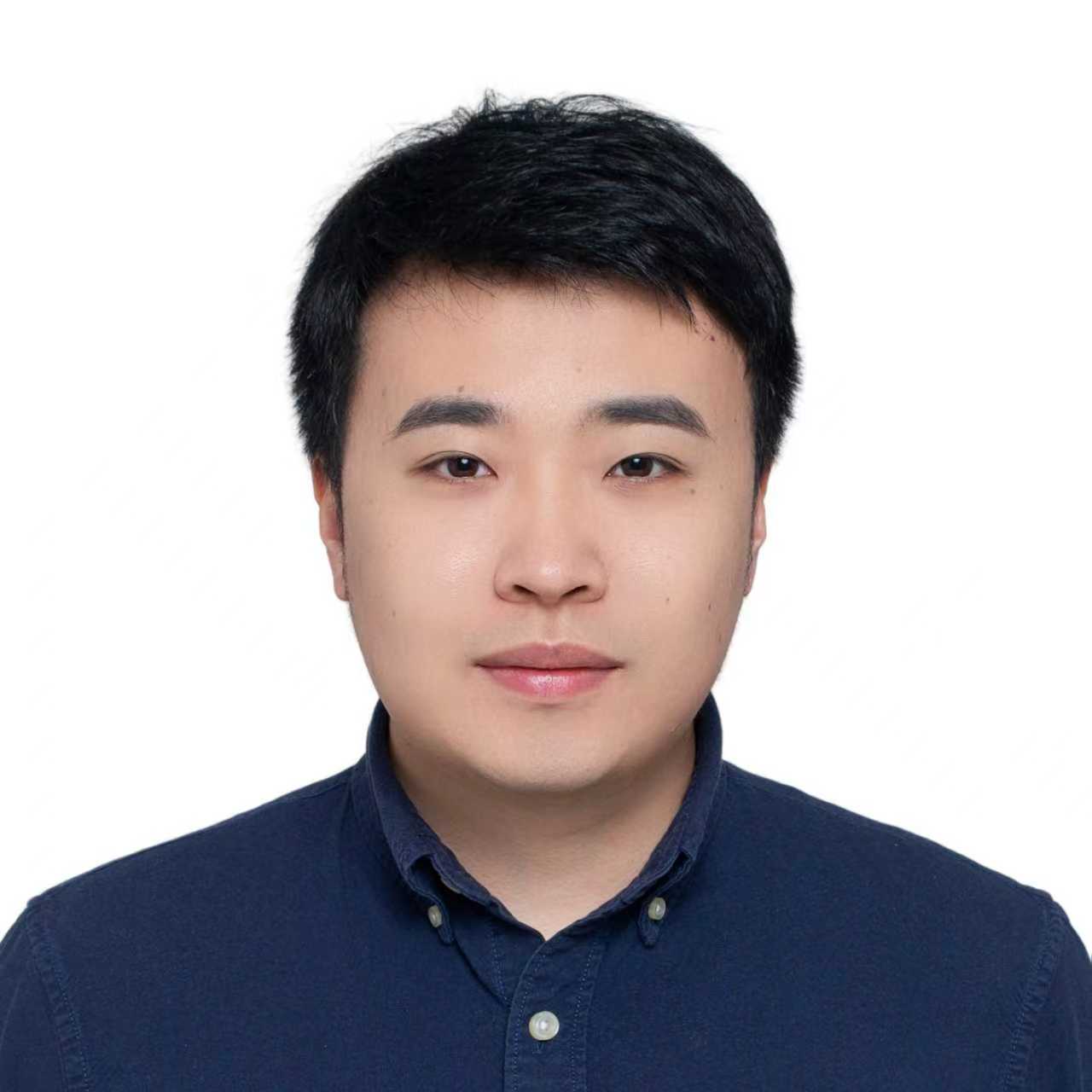}}]{Xinyang Liu}
received the B.S. and M.S. degrees in electronic engineering from Xidian University, Xi’an, China, in 2021 and 2024, respectively. He is currently a Ph.D. student at The University of Texas at Austin, Austin, Texas, USA. His research focuses on generative modeling, including theoretical foundations and a wide range of applications in data generation and multimodal learning.
\end{IEEEbiography}
\vspace{-5mm}
\begin{IEEEbiography}[{\includegraphics[width=1in,height=1.25in,clip,keepaspectratio]{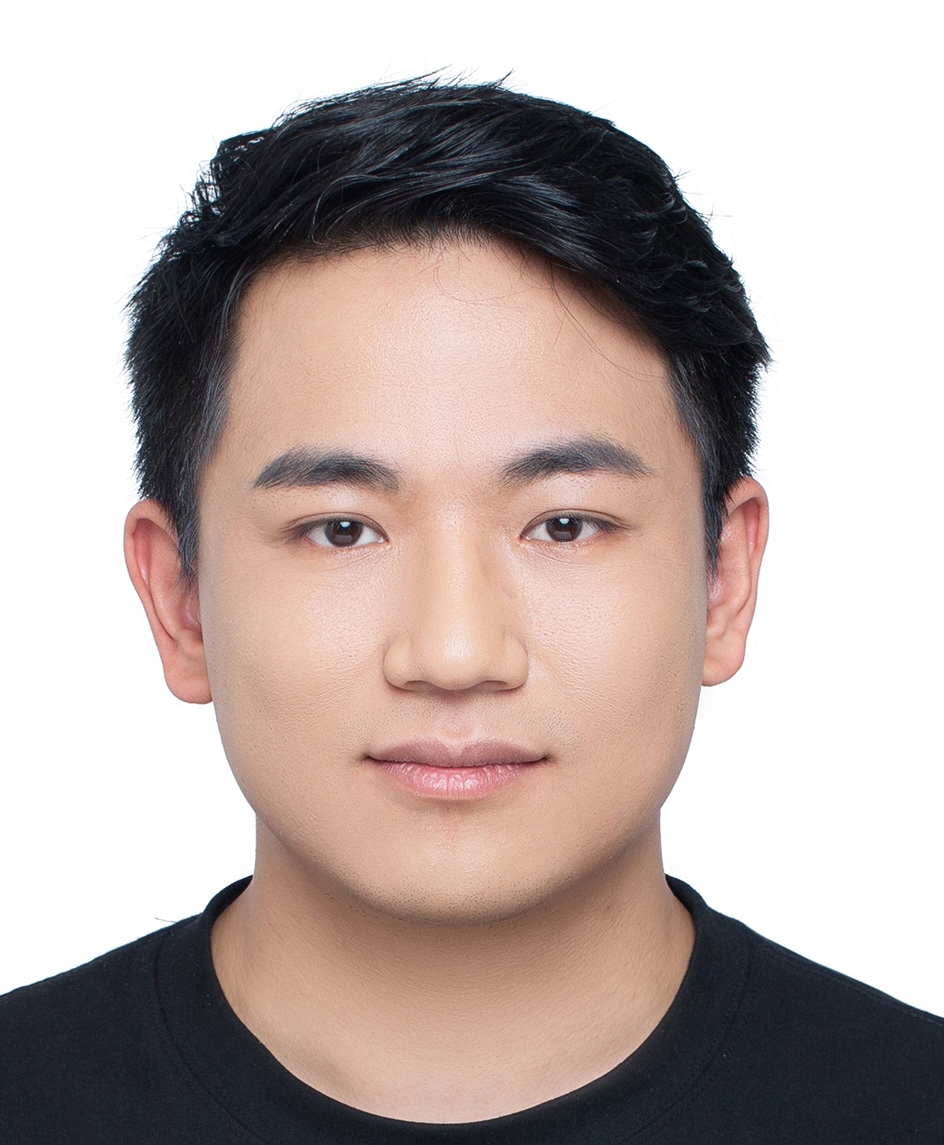}}]{Yuxin Li} received the B.S. degree in electronic information science and technology from Hefei University of Technology, Hefei, China, in 2018, and the M.S. degree in weapon systems and engineering applications from Xi’an Electronic Engineering Research Institute, Xi’an, China, in 2021. He received the Ph.D. degree in information and communication engineering from Xidian University, Xi’an, China, in 2025. He is currently a Postdoctoral Researcher with the National Key Laboratory of Radar Signal Processing, Xidian University, Xi’an, China. His research interests include radar signal processing, statistical machine learning, probabilistic modeling, and automatic target recognition.
\end{IEEEbiography}
\vspace{-5mm}
\begin{IEEEbiography}[{\includegraphics[width=1in,height=1.25in,clip,keepaspectratio]{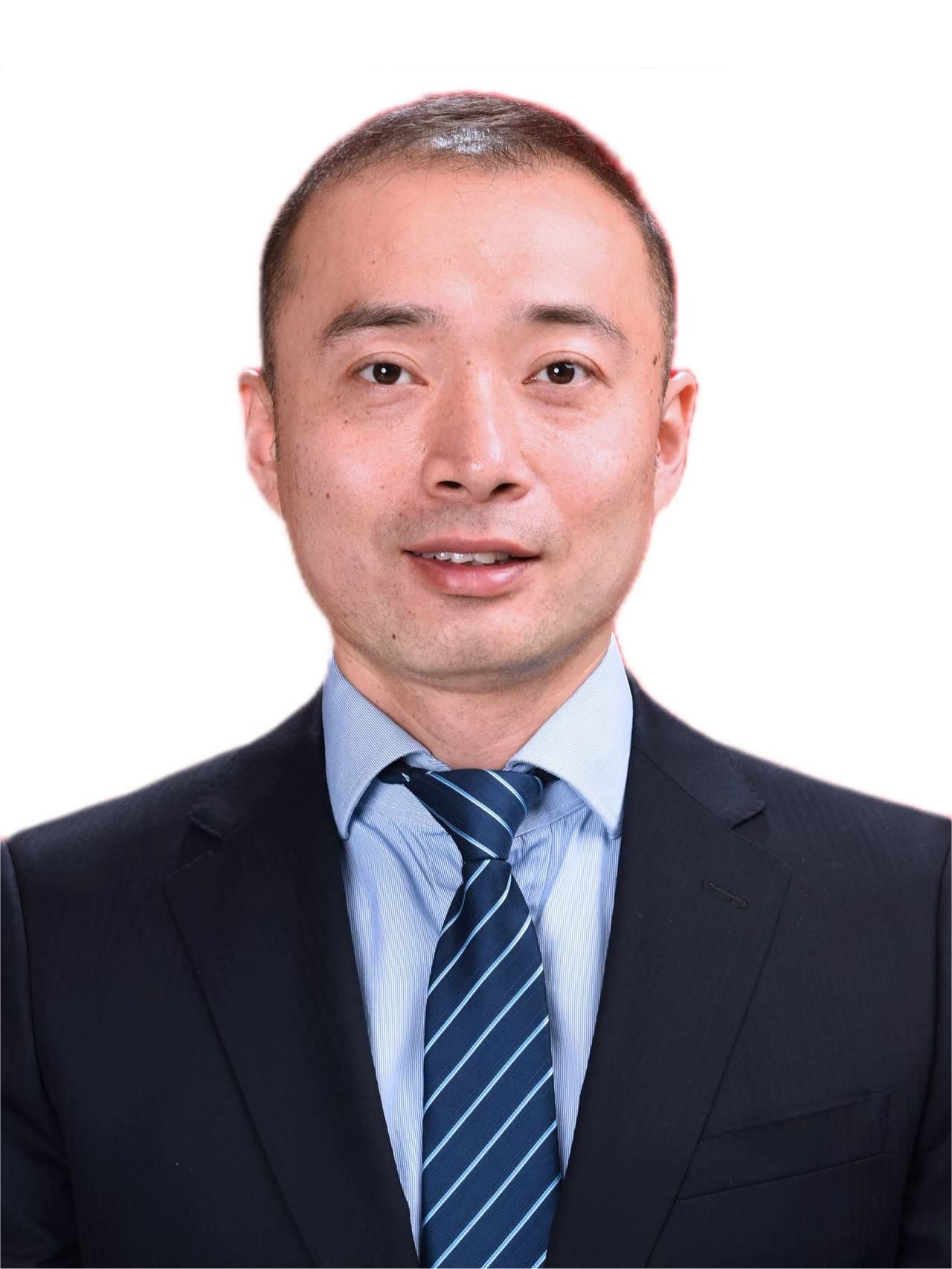}}]{Bo Chen}
	(Senior Member, IEEE) received the BS,
	MS, and PhD degrees in electronic engineering from
	Xidian University, Xi’an, China, in 2003, 2006, and
	2008, respectively. He became a postdoctoral Fellow, a Research Scientist, and a Senior Research
	Scientist with the Department of Electrical and Computer Engineering, Duke University, Durham, NC,
	USA, from 2008 to 2012. From 2013, he has been a
	Professor with the National Laboratory for Radar
	Signal Processing, Xidian University. He was the
	recipient of the Honorable Mention for 2010 National
	Excellent Doctoral Dissertation Award and is selected into Overseas Talent
	by Chinese Central Government, in 2014. His current research interests
	include statistical machine learning, statistical signal processing, and radar
	automatic target detection and recognition.
\end{IEEEbiography}
\vspace{-10mm}
\begin{IEEEbiography}[{\includegraphics[width=1in,height=1.25in,clip,keepaspectratio]{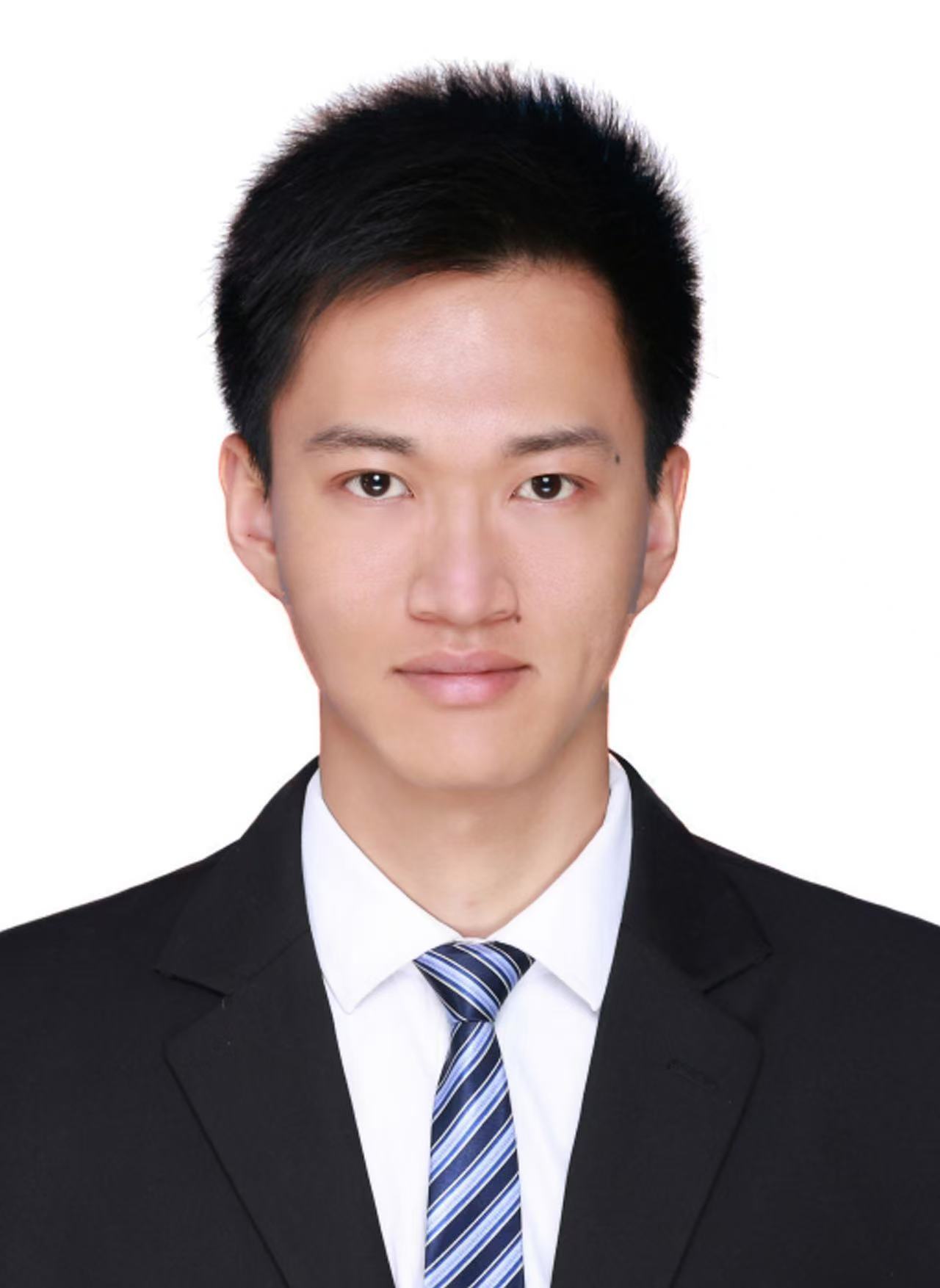}}]{Chaojie Wang}
	received the BS and
	PhD degrees in electronic engineering from Xidian
	University, Xi’an, China, in 2016 and 2021, respectively. 
	He is currently working as a research Fellow
	with Nanyang Technological University, Singapore.
	His research interests focus on statistical machine
	learning and its combinations with real-world applications, including multimodal learning, natural language processing, and knowledge graphs.
\end{IEEEbiography}
\vspace{-10mm}
\begin{IEEEbiography}[{\includegraphics[width=1in,height=1.25in,clip,keepaspectratio]{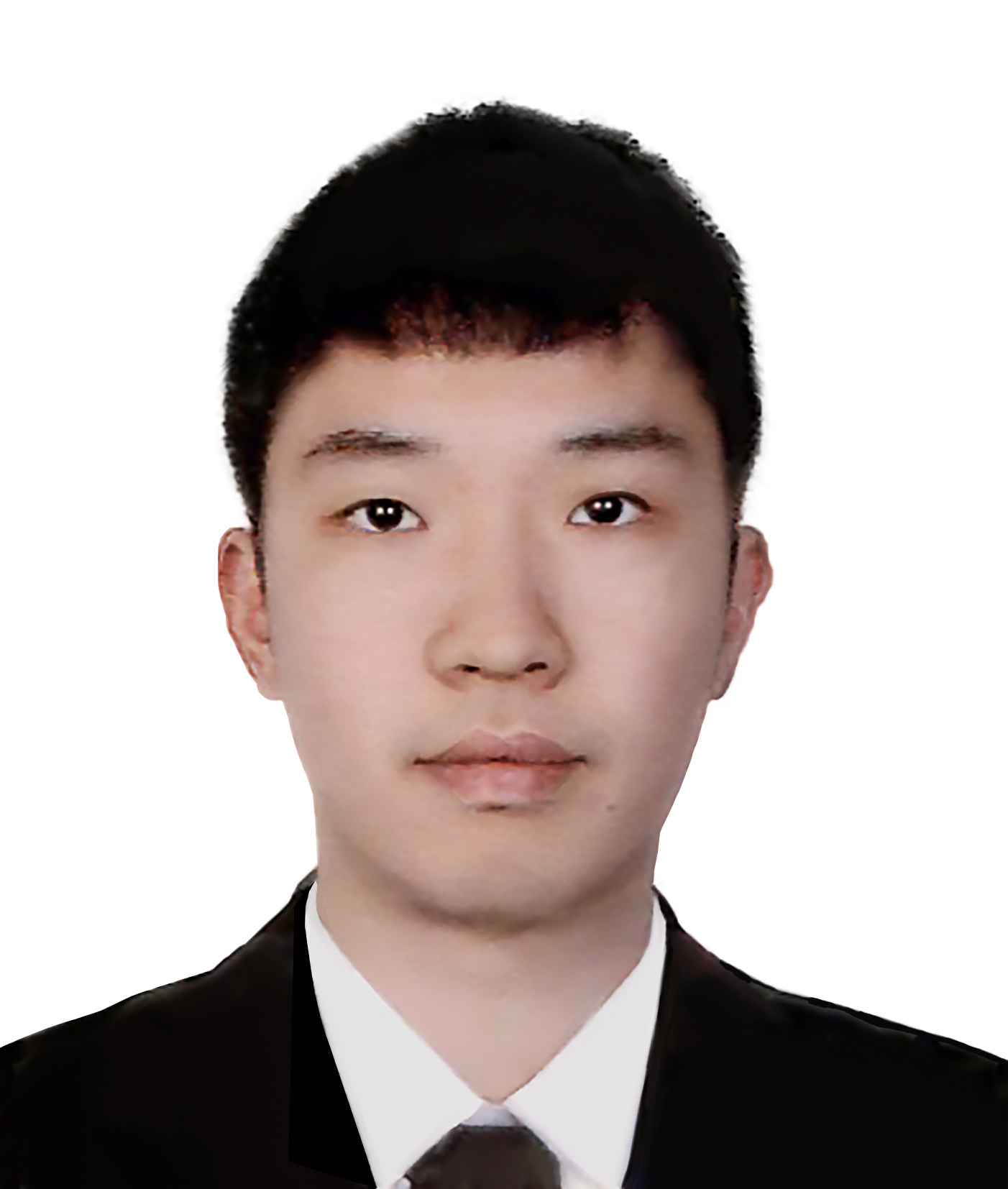}}]{Yilin He}
	received the B.S. degree in Information Science and Technology from the University of Science and Technology of China, and the Ph.D. degree in Statistics from The University of Texas at Austin. His research interests include Bayesian machine learning, graph representation learning, and generative models.
\end{IEEEbiography}
\vspace{-10mm}
	\begin{IEEEbiography}[{\includegraphics[width=0.95in,height=1.25in,clip,keepaspectratio]{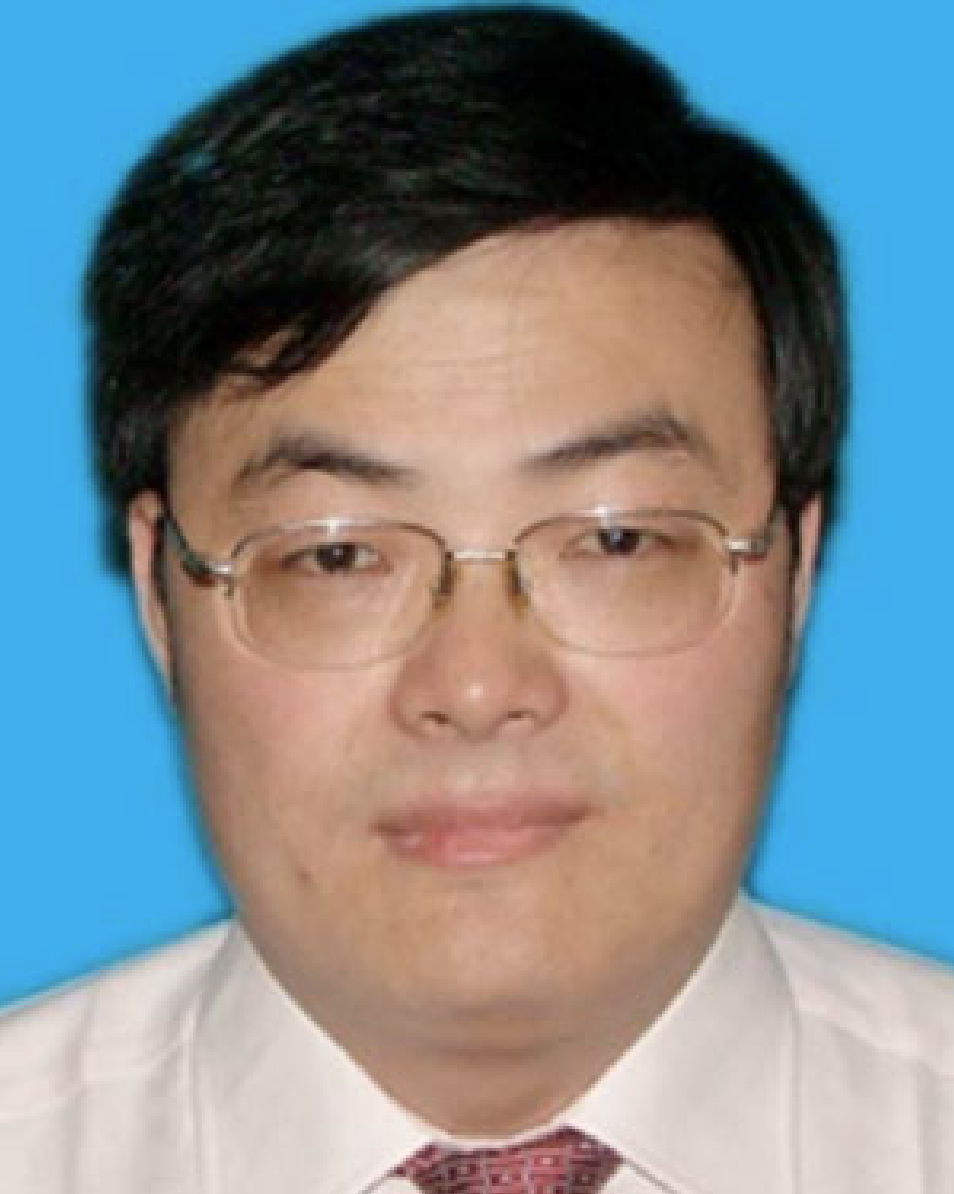}}]{Hongwei Liu}(Member, IEEE) received the M.S. and Ph.D. degrees in electronic engineering from Xidian University, Xi’an, China, in 1995 and 1999, respectively. From 2001 to 2002, he was a Visiting Scholar with the Department of Electrical and Computer Engineering, Duke University, Durham, NC, USA. He is currently a Professor with the National Laboratory of Radar Signal Processing, Xidian University. His research interests include radar automatic target recognition, radar signal processing, and adaptive signal processing.
\end{IEEEbiography}
\vspace{-10mm}
\begin{IEEEbiography}[{\includegraphics[width=1.0in,height=1.25in,clip,keepaspectratio]{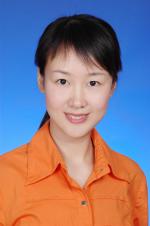}}]{Xuefei Cao} (Member, IEEE) received the B.Sc.
	degree in telecommunication engineering and the
	M.Sc. and Ph.D. degrees in communication and
	information systems from Xidian University, Xi’an,
	China, in 2003, 2006, and 2008, respectively.
	From 2010 to 2012, she worked as a Postdoctoral
	Researcher with Duke University, Durham, NC,
	USA. She is the first author of about 20 research
	papers, including papers published in the IEEE
	TRANSACTIONS ON VEHICULAR TECHNOLOGY
	and Information Sciences. One of her research
	results has been cited more than 240 times. Her research interests include
	information security, communication network security, network intrusion
	detection, and machine learning.
\end{IEEEbiography}
\vspace{-10mm}
\begin{IEEEbiography}[{\includegraphics[width=1in,height=1.25in,clip,keepaspectratio]{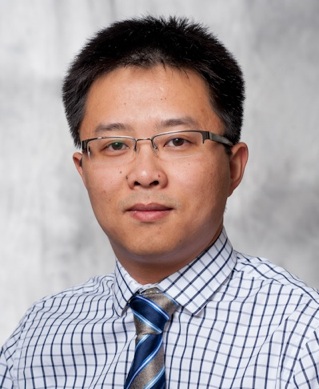}}]{Mingyuan Zhou}received the BS degree in acoustics with the Department of Electronic Science and
	Engineering from Nanjing University, Nanjing,
	China, in 2005, the MS degree in signal and information processing from Chinese Academy of Sciences, Beijing, China, in 2008, and the PhD degree
	in electrical and computer engineering from Duke
	University, Durham, NC, USA, in 2013. 
	He is a
	full professor of Statistics with the McCombs
	School of Business, University of Texas with Austin,
	Austin, TX, USA. His research interest lies at the
	intersection of Bayesian statistics and machine learning, covering a diverse
	set of research topics in statistical theory and methods, hierarchical models, Bayesian nonparametrics, statistical inference for big data, and deep
	learning. He is currently focused on advancing both statistical inference
	with deep learning and deep learning with probabilistic methods. He has
	served as the 2018–2019 Treasurer of the Bayesian Nonparametrics Section, International Society for Bayesian Analysis, and as area Chairs for
	leading machine learning conferences, including NeurIPS 2017–2025,
	ICLR 2019-2025, and AAAI 2020.
\end{IEEEbiography}

\end{document}